\documentclass{article}

 \usepackage[main, final]{neurips_2026}

\usepackage[utf8]{inputenc} 
\usepackage[T1]{fontenc}    
\usepackage{hyperref}       
\usepackage{url}            
\usepackage{booktabs}       
\usepackage{amsfonts}       
\usepackage{nicefrac}       
\usepackage{microtype}      
\usepackage{xcolor}         
\usepackage{hyperref}       
\usepackage{url}            
\usepackage{amsfonts}       
\usepackage{nicefrac}       
\usepackage{microtype}      
\usepackage{xcolor}         
\usepackage{amsmath} 
\usepackage{makecell}
\usepackage{hyperref}       
\usepackage{url}            
\usepackage{amsfonts}       
\usepackage{nicefrac}       
\usepackage{microtype}      
\usepackage{xcolor}         
\usepackage{amsmath}
\usepackage{amssymb}
\usepackage{booktabs}
\usepackage{bbm}
\usepackage{pifont}        
\usepackage{xcolor}        
\usepackage{multirow}
\usepackage{graphicx}
\usepackage{subcaption}
\usepackage{float}
\usepackage[inline]{enumitem}
\usepackage{subcaption}
\usepackage{wrapfig}
\usepackage{titletoc}
\usepackage{etoc}
\usepackage{algorithm}
\usepackage{algpseudocode}
\newcommand{\cmark}{\textcolor{green!60!black}{\ding{51}}} 
\newcommand{\xmark}{\textcolor{red}{\ding{55}}}            

\title{StepCAD: Mesh-to-CAD Code Generation via LLM Policy and Geometry-Guided Search}

\author{%
  Ghadi Nehme \\
  Department of Mechanical Engineering\\
  Massachusetts Institute of Technology\\
  Cambridge, MA 02139\\
  \texttt{ghadi@mit.edu}\\
  \And
  Faez Ahmed \\
  Department of Mechanical Engineering\\
  Massachusetts Institute of Technology\\
  Cambridge, MA 02139\\
  \texttt{faez@mit.edu}\\
}

\begin{document}

\maketitle

\begin{abstract}
Recovering executable CAD programs from 3D meshes is challenging due to the compositional nature of CAD construction and the interaction between discrete modeling choices and continuous parameters. Many learning-based methods predict complete programs in a single pass and rely predominantly on sketch-extrude representations, limiting operation diversity and opportunities to correct geometric errors during reconstruction. We introduce StepCAD, a generative optimization approach that combines a state-conditioned CAD policy with geometry-guided search. Given an input mesh, the policy predicts construction actions conditioned on both target and intermediate geometry, and an IoU-guided tree search refines the resulting program through local edits. We also introduce ARCADE-1.5M, a large-scale dataset of 1.5M executable CAD programs spanning diverse operations, sequences with a maximum length of 150+ counted operations, and 12.5M intermediate state-action transitions. Experiments across multiple CAD reconstruction benchmarks show that StepCAD achieves state-of-the-art geometric reconstruction accuracy with consistently high validity, yielding up to 87.2\% relative IoU improvement over the strongest evaluated baseline, with particularly large gains on complex shapes. Project page: \url{https://ghadinehme.com/stepcad.github.io/}
\end{abstract}

\section{Introduction}
Recovering structured, executable CAD programs from 3D meshes is a critical capability for modern engineering workflows. Many industrial pipelines produce dense, unstructured 3D geometry — from 3D scans of physical parts~\cite{karadeniz2025micadangelo,rukhovich2025cad}, topology-optimized designs~\cite{ren2025amrto,amroune2022automated}, and increasingly from generative 3D models~\cite{xiang2025structured,zhao2025hunyuan3d,li2025caddreamer} — that must be converted into editable parametric representations to enable design iteration, manufacturing, and downstream simulation. Our goal is geometry-driven reverse engineering: recovering an executable construction sequence that reproduces the input, without claiming recovery of its original design intent or constraints. Unlike conventional 3D reconstruction, which approximates surface geometry, \emph{mesh-to-CAD generation} seeks to recover a sequence of parametric operations that reconstructs the shape compactly and faithfully. This task is inherently challenging: CAD programs are long-horizon, compositional, and highly structured, with strong dependencies between operations and continuous parameters that must be estimated with high precision~\cite{wu2021deepcad,xu2023hierarchical}.

Recent learning-based approaches model CAD reconstruction as autoregressive program generation conditioned on geometry~\cite{wu2021deepcad,rukhovich2025cad,kolodiazhnyi2025Cadrille,xu2024cad}. Early approaches and datasets emphasize sketch-and-extrude pipelines, while recent procedural datasets expand the use of standard CAD operations~\cite{ataei2026zero,elistratov2026CADEvolve}. Nevertheless, predicting a complete program without executing intermediate states limits opportunities to respond to the evolving reconstruction. Learned refinement methods introduce feedback, but geometric errors can remain in both the program structure and its continuous parameters. This motivates combining state-conditioned generation with explicit, executable program-space search.

We argue that mesh-to-CAD reconstruction should be reformulated as policy-driven, geometry-conditioned program synthesis with search-based refinement. This perspective decomposes the problem into two complementary components: a learned policy that conditions on intermediate geometry to propose candidate construction sequences, and a geometry-aware optimization procedure that refines these sequences to improve geometric fidelity. The formulation leverages the strengths of generative models for capturing global structure while correcting geometric discrepancies through explicit search guided by geometric objectives — a paradigm with a long history in program synthesis~\cite{balog2016deepcoder,ellis2019write,chen2018execution} that we adapt to the unique constraints of CAD generation.

Realizing this paradigm requires both a model capable of state-conditioned generation and training data that exposes diverse construction trajectories — neither of which is well-supported by existing CAD datasets. We therefore introduce two contributions in tandem. First, ARCADE-1.5M, a large-scale synthetic dataset of 1.5M executable CAD programs spanning extrude, revolve, loft, sweep, Boolean operations, and finishing features, with sequences reaching 150+ counted operations and 12.5M intermediate state-action transitions providing dense supervision over construction trajectories. Second, StepCAD, a generative optimization~\cite{regenwetter2024generative} approach for mesh-to-CAD reconstruction that, given an input shape, generates a candidate program using a state-conditioned LLM policy trained on ARCADE-1.5M, then refines it through a geometry-aware tree search over local edits — combining continuous parameter optimization with discrete structural modifications, all guided by IoU.

This combination of policy and optimization yields substantial improvements in reconstruction accuracy and robustness. We show that explicit refinement of program structure and numerical parameters improves geometric fidelity beyond generation alone. Here, geometric fidelity refers to agreement with the target shape, not recovery of every parameter in a reference program. StepCAD achieves state-of-the-art (SOTA) performance on CADBench \cite{doris2026cadbench},  a CAD benchmark spanning ABC \cite{koch2019abc}, Fusion 360 \cite{willis2021fusion}, DeepCAD \cite{wu2021deepcad}, MCB \cite{kim2020large}, and Objaverse \cite{deitke2023objaverse}, with increasingly larger gains as shape complexity grows, precisely where prior approaches degrade most severely.

In summary, our contributions are:
\begin{itemize}[leftmargin=1.2em]
\item \textbf{ARCADE-1.5M: a large-scale CAD program dataset.} We introduce a dataset of 1.5M executable CAD programs with rich operation diversity, long-horizon construction sequences (150+ counted operations), and 12.5M intermediate state-action transitions, substantially expanding the scale and complexity of available CAD data.
\item \textbf{State-conditioned CAD policy.} We propose a point-cloud-conditioned autoregressive policy that predicts CAD operations given both intermediate and target geometry.
\item \textbf{Program-space optimization for CAD generation.} We reformulate mesh-to-CAD reconstruction as optimization over program space and introduce a geometry-aware tree search that refines candidate programs through IoU-guided local edits.
\item \textbf{State-of-the-art performance on diverse CAD benchmarks.} We achieve SOTA results across five diverse benchmarks, with up to \textbf{87.2\%} relative improvement in IoU over the best prior method, demonstrating strong robustness and growing advantages as shapes become more complex.
\end{itemize}

\section{Related Work}

\textbf{Learning-Based CAD Program Generation.}
While text-to-CAD generation has attracted significant attention~\cite{he2025cad,wang2025text,khan2024text2cad}, industrial workflows more often require recovering CAD programs from existing 3D geometry — such as scanned parts~\cite{karadeniz2025micadangelo,rukhovich2025cad,liu2024point2cad}, topology-optimized geometry~\cite{ren2025amrto,amroune2022automated,larsen2009converting}, or AI-generated meshes~\cite{xiang2025structured,zhao2025hunyuan3d,li2025caddreamer} — where the design already exists in shape but lacks editable parametric structure. Recent work has therefore explored neural approaches for generating CAD programs as sequences of parametric operations from 3D shapes~\cite{rukhovich2025cad,xu2023hierarchical,kolodiazhnyi2025Cadrille}, motivated by applications in reverse engineering, scalable dataset construction, and bridging geometry-centric 3D generation~\cite{xiang2025structured,zhao2025hunyuan3d} with editable, parametric workflows. Early methods such as DeepCAD~\cite{wu2021deepcad} introduce domain-specific languages for sketch-and-extrude programs and learn sequence autoencoders over this representation, with subsequent works extending the paradigm to richer inputs and structured decoders~\cite{yu2025gencad,xu2022skexgen,xu2023hierarchical, man2026videocad}. More recently, large language models have been applied to CAD program generation from images~\cite{li2025caddreamer,chen2025cadcrafter,doris2025cad} and point clouds~\cite{rukhovich2025cad,kolodiazhnyi2025Cadrille,xu2024cad}, enabling flexible conditioning across modalities. Despite these advances, existing methods rely primarily on amortized autoregressive prediction, which can leave geometric discrepancies in the resulting shape. In contrast, StepCAD explicitly separates program generation from program optimization, using a learned policy to propose structurally consistent programs and a search procedure to refine them for geometric accuracy.

\textbf{Iterative Refinement and Neural-Guided Program Search.}
A small but growing line of work explores iterative refinement for CAD generation. CADReasoner~\cite{kabisov2026cadreasoner} introduces LLM-based refinement guided by visual feedback, CADCrafter~\cite{chen2025cadcrafter} applies direct preference optimization with a code-validity checker, and CADFusion~\cite{wang2025text} alternates supervised and visual-feedback stages during training. These methods incorporate feedback through different training or generation procedures. StepCAD instead makes execution-scored local program edits an explicit inference-time optimization stage. More broadly, combining learned policies with explicit search has a long history in program synthesis~\cite{balog2016deepcoder,ellis2019write,chen2018execution} and recent code generation~\cite{ni2023lever,chen2023teaching,zhang2023planning}, where neural networks propose candidates that are refined by search to satisfy correctness criteria. StepCAD adopts this paradigm in the CAD setting: an LLM policy proposes structurally consistent programs, while a geometry-aware tree search refines them through local edits — including continuous parameter optimization, sketch modification, and operation skipping — with candidates ranked by IoU against the target. Compared to general code generation, CAD program search is uniquely constrained by both executability and continuous parameter precision, motivating geometry as the primary search signal.

\textbf{Entity-grounded and feedback-driven CAD generation.}
Pointer-CAD~\cite{qi2026pointer} conditions on text and intermediate B-reps and selects referenced edges and faces through a pointer mechanism. FutureCAD~\cite{li2026towards} grounds textual selection queries to B-rep primitives. Seek-CAD~\cite{li2026seek} uses visual and chain-of-thought feedback for iterative generation. Our focus is mesh-conditioned reconstruction, where an explicit target shape enables direct geometric scoring of executable program edits. StepCAD combines this objective with state-conditioned proposals and a refinement stage that can also process compatible programs from other generators. Its coarse selectors do not provide the fine-grained entity addressing of Pointer-CAD or FutureCAD. Contemporaneous Pointer-CAD v2~\cite{qi2026pointerv2} further motivates distinguishing geometric overlap from dimension-aware and entity-level accuracy.

\textbf{Geometry-Driven CAD Reverse Engineering.}
Another line of work recovers CAD structure directly from 3D shape \cite{nehme2026cadfit}. Classical pipelines decompose shapes into primitives and fit analytic surfaces but struggle to produce consistent parametric programs for complex designs. Learning-based approaches improve robustness by combining neural segmentation with surface fitting, particularly for boundary representation (B-rep) reconstruction~\cite{liu2024point2cad,xu2024brepgen,lee2025brepdiff,uy2022point2cyl,karadeniz2025micadangelo}. B-rep reconstruction alone does not recover an ordered feature history; recovering geometry or an executable history also does not establish recovery of design intent. Inverse constructive solid geometry (CSG) methods~\cite{du2018inversecsg} formulate reconstruction as a combinatorial search over Boolean programs, but are typically restricted to simple primitive sets and do not scale to sketch-based modeling, revolutions, or finishing features. In contrast, StepCAD explicitly models the sequential construction process and produces executable programs, enabling both interpretability and editability.

\textbf{CAD Datasets.}
Large-scale datasets such as ABC~\cite{koch2019abc}, DeepCAD~\cite{wu2021deepcad}, and Fusion360~\cite{willis2021fusion} have enabled learning structured CAD representations. Their supervision differs: DeepCAD and the Fusion 360 Gallery construction task emphasize sketch-and-extrude sequences, whereas ABC supplies geometry without an ordered construction history. Concurrent efforts such as CAD-Recode~\cite{rukhovich2025cad}, Zero-to-CAD~\cite{ataei2026zero}, and CADEvolve~\cite{elistratov2026CADEvolve} expand operation diversity through procedural and evolutionary synthesis, using many of the same standard CadQuery operations as our corpus. Our ARCADE-1.5M dataset is designed to provide long-horizon, diverse, and executable CAD programs with dense intermediate supervision, enabling effective training of state-conditioned policies.

\section{Method}

We hypothesize that learning CAD programs from geometry requires large-scale procedural data with rich operation diversity, which is lacking in existing datasets. To address this, we introduce \textbf{ARCADE (Artificial Repository of CAD Examples)}, a large-scale synthetic dataset of executable CAD programs with diverse operations and workflow-consistent structure, providing dense supervision over construction trajectories. We then propose \textbf{StepCAD}, a generative optimization approach for mesh-to-CAD reconstruction: a \textbf{state-conditioned LLM policy} generates candidate programs, followed by \textbf{geometry-aware optimization} in program space. This decomposition enables learning global construction structure while achieving high geometric fidelity.

\subsection{Problem Statement}
\label{sec:problem}

We study \textbf{mesh-to-CAD reconstruction}: recovering an executable parametric CAD program that reproduces a target 3D shape. Let $\mathcal{M}$ denote the space of meshes and $\mathcal{P}$ the space of valid CAD programs. A program $p = (a_1,\dots,a_T) \in \mathcal{P}$ is an ordered sequence of construction actions whose execution produces geometry $S(p)\in\mathcal{M}$. Given an input mesh $m \in \mathcal{M}$, the goal is to predict a valid program $\hat p$ such that the generated geometry matches the target shape: $\hat p = \arg\max_{p\in\mathcal{P}} \mathrm{IoU}(S(p), m)$. The problem is challenging due to the combinatorial structure of $\mathcal{P}$, long-horizon autoregressive dependencies, and the existence of multiple valid programs producing similar geometry. We represent CAD programs using CadQuery code \cite{cadquery_contributors_2025_14590990}.

\begin{figure}[t]
    \centering
    \includegraphics[width=1\linewidth]{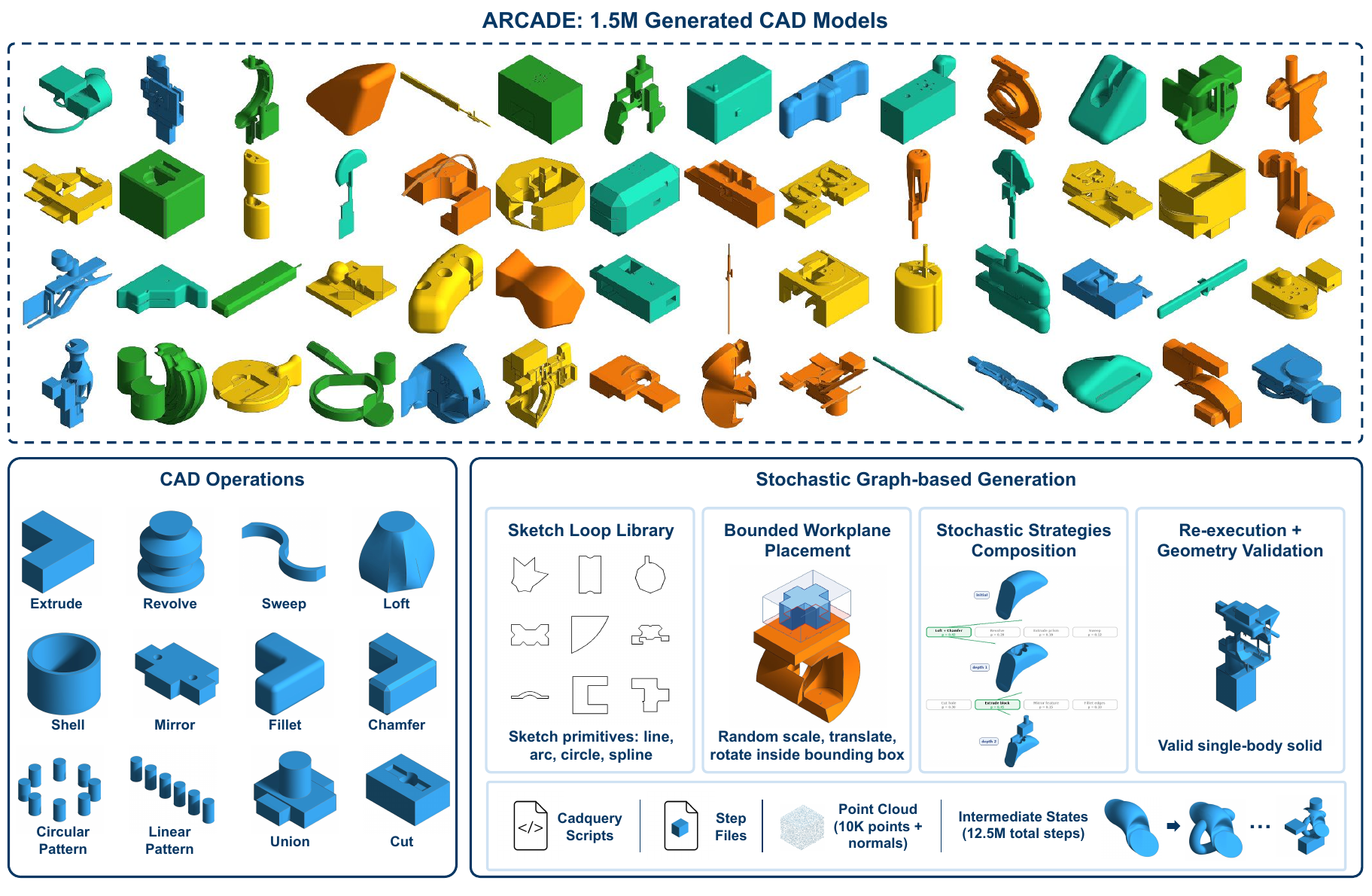}
    \caption{
    \textbf{ARCADE-1.5M: large-scale synthetic CAD program generation.}
    Our pipeline synthesizes diverse long-horizon CAD programs through stochastic strategy-guided composition of sketch primitives and CAD operations. Each program is executed and validated, yielding 1.5M executable programs and 12.5M intermediate construction states with broad operation diversity.
    }
    \label{fig:arcade}
\end{figure}

\subsection{ARCADE-1.5M: Constraint-Guided Synthetic CAD Programs}
\label{sec:ARCADE-1.5M}

We introduce \textbf{ARCADE-1.5M}, a dataset of \textbf{1.5M} executable parametric CAD programs paired with valid 3D geometry. ARCADE-1.5M combines stochastic geometry generation with workflow-inspired constraints to synthesize long-horizon, structurally plausible CAD construction sequences at scale (Fig.~\ref{fig:arcade}). Dataset statistics and generation details can be found in Appendices \ref{app:dataset-stats} and \ref{app:dataset}.

\textbf{Constraint-guided primitive generation.}
We initialize generation from a library of 2D sketch loops extracted from ABC, including lines, arcs, circles, and splines. Sketches are randomly perturbed, rotated, and scaled, then placed on randomly selected faces of a target bounding box, providing explicit control over feature size, placement, and orientation. Each sketch is converted into a solid using randomly sampled \textbf{extrude}, \textbf{revolve}, \textbf{loft}, or \textbf{sweep} operations with stochastic parameters. Optional fillets and chamfers are applied probabilistically to increase geometric diversity.

\textbf{Workflow-inspired composition constraints.}
Rather than composing disconnected primitives, ARCADE-1.5M enforces constraints inspired by practical CAD workflows. All intermediate states must remain valid single-body solids with positive volume. Additive operations require controlled overlap with the host body, while subtractive operations are accepted only under sufficient volumetric intersection. These constraints prevent degenerate geometry and preserve physically meaningful feature placement.

\textbf{Strategy-based hierarchical construction.}
To generate long, coherent programs, ARCADE-1.5M uses \textbf{94 heuristic generation strategies} that constrain operation types and approximate sequence lengths. Strategies emulate common modeling patterns, including additive construction, cut-heavy machining, shelling, mirroring, lofting, sweeping, and finishing operations. This hierarchical structure enables realistic long-horizon programs while avoiding combinatorial explosion. Generated sequences range from 3 to 150+ operations (mean: 40.3, median: 33).

\textbf{Operations, actions, and API calls.}
An \emph{operation} in our dataset statistics is a sketch primitive, solid-modeling operation, or Boolean combination. A \emph{solid operation} is an extrude, revolve, sweep, loft, fillet, chamfer, shell, or mirror. A policy \emph{action} bundles a solid operation with its associated Boolean combination and, for sketch-based operations, the profile-defining sketch primitives. Finishing operations do not require a new sketch. Thus, the mean of 40.3 counted operations per program differs from the mean of 8.5 solid operations (median 8), corresponding to approximately $40.3/8.5=4.7$ counted operations per construction action. An API call is a code-level method invocation; setup, selection, and export calls should not be confused with the operation-count categories. Actions are defined at geometry-update boundaries so that the next policy step can condition on a new solid.

\textbf{Validation and normalization.}
Every sample undergoes double validation: the in-memory CadQuery solid must be valid, and the exported Python program is re-executed in a clean namespace before STEP export. Shapes are normalized to a canonical $[-1,1]^3$ bounding box, and floating-point literals are rounded to three decimals to reduce token complexity.

\textbf{Dataset characteristics.}
Each sample contains a CadQuery script, STEP B-Rep, triangulated mesh, and a 10K-point cloud with normals. We additionally store intermediate geometry after every \textbf{solid operation}, yielding approximately \textbf{12.5M} construction states for dense trajectory supervision. ARCADE-1.5M covers workplanes, sketch primitives, volumetric and Boolean operations, shells, mirrors, fillets, chamfers, and spatial patterns. Its contribution is the distribution and composition of these standard operations, together with scale and intermediate supervision, rather than a new operation vocabulary. For example, sweep occurs in 25.37\% of programs, compared with 3.88\% for Zero-to-CAD and 5.75\% for CADEvolve; the mean face count is 86 (Table~\ref{tab:cad_datasets}). Overall, ARCADE-1.5M bridges executable CAD construction logic and resulting 3D geometry, enabling training for realistic mesh-to-CAD generation.

\begin{table*}[t]
\centering
\small
\setlength{\tabcolsep}{3.5pt}
\renewcommand{\arraystretch}{1.15}
\resizebox{\textwidth}{!}{%
\begin{tabular}{ccccccccccccccc}
\toprule
\textbf{Dataset} & \textbf{Size} & \makecell{\textbf{Has Construction} \\ \textbf{Sequence}} & \makecell{\textbf{Avg} \\ \textbf{Ops / Seq}} & \makecell{\textbf{Avg Solid} \\ \textbf{Ops / Seq}} & \makecell{\textbf{Avg} \\ \textbf{Faces}}
& \multicolumn{8}{c}{\textbf{Solid Operation Coverage (\%)}} \\
\cmidrule(lr){7-14}
& & & & & 
& \textbf{Extrude} & \textbf{Revolve} & \textbf{Sweep} & \textbf{Loft} & \textbf{Fillet} & \textbf{Chamfer} & \textbf{Shell} & \textbf{Mirror} \\
\midrule

DeepCAD \cite{wu2021deepcad} 
& 178k & \cmark & 15 & 1.5 & 10 
& 100 & 0 & 0 & 0 & 0 & 0 & 0 & 0 \\

ABC-mono \cite{chen2025img2cad} 
& 200k & \cmark & * & * & * 
& 100 & 0 & 0 & 0 & 0 & 0 & 0 & 0 \\

KOCAD \cite{chen2025img2cad} 
& 4.6k & \cmark & * & * & * 
& 100 & 0 & 0 & 0 & 0 & 0 & 0 & 0 \\

Fusion 360 \cite{willis2021fusion}  
& 8.6k & \cmark & 18 & 2.2 & 16 
& 100 & 0 & 0 & 0 & 0 & 0 & 0 & 0 \\

CAD-Recode \cite{rukhovich2025cad} 
& 1M & \cmark & 28 & 2.0 & 17 
& 100 & 0 & 0 & 0 & 0 & 0 & 0 & 0 \\

Zero-to-CAD \cite{ataei2026zero} 
& 1M & \cmark & 24 & 5.5 & 44 
& \textbf{85.99} & \textbf{18.73} & 3.88 & 4.17 & \textbf{36.92} & \textbf{74.66} & \textbf{15.00} & 1.56 \\

CADEvolve \cite{elistratov2026CADEvolve} 
& 1.3M & \cmark & 28 & 5.5 & 54 
& \textbf{83.05} & 4.80 & \textbf{5.75} & \textbf{8.48} & \textbf{27.78} & 4.76 & 1.95 & 0.08 \\

ABC \cite{koch2019abc} 
& 1M & \xmark & -- & -- & 114
& -- & -- & -- & -- & -- & -- & -- & -- \\

\midrule
\textbf{ARCADE-1.5M (ours)} 
& \textbf{1.5M} & \cmark & \textbf{40} & \textbf{8.5} & \textbf{86}
& \textbf{95.42} & \textbf{12.13} & \textbf{25.37} & \textbf{13.10} & \textbf{34.77} & \textbf{33.66} & \textbf{10.01} & \textbf{6.70} \\

\bottomrule
\end{tabular}
}
\caption{\textbf{Comparison of CAD datasets with operation coverage.}
Operation coverage is the percentage of programs containing at least one instance of the operation. Prior datasets are mostly sketch-extrude only. Concurrent works CADEvolve and Zero-to-CAD add parametric diversity, while ARCADE-1.5M has the most programs, highest mean operation counts, and highest mean face count among the listed datasets with construction
sequences. 
$*$: datasets that are not publicly available.}
\label{tab:cad_datasets}
\end{table*}

\subsection{StepCAD: Generative Optimization for Mesh-to-CAD Reconstruction}

\begin{figure}
\centering
\includegraphics[width=1\linewidth]{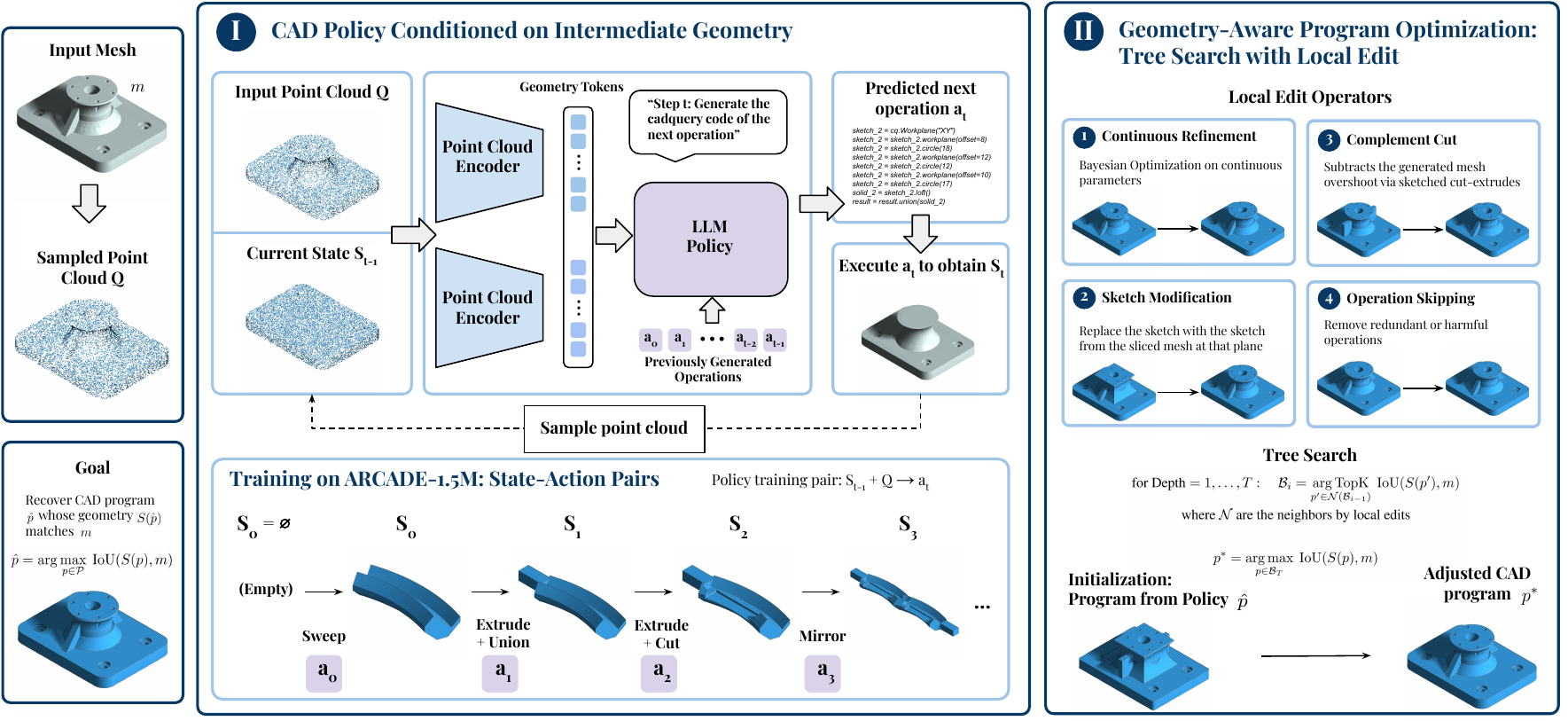}
\caption{
\textbf{StepCAD pipeline.}
A state-conditioned CAD policy generates an executable program from an input shape, which is subsequently refined via geometry-aware search over local program edits.
}
\label{fig:pipeline}
\end{figure}

We propose \textbf{StepCAD}, a two-stage framework that combines a \emph{state-conditioned autoregressive CAD policy} ($\Pi$-StepCAD) with \emph{geometry-aware program optimization} (Fig.~\ref{fig:pipeline}). The policy generates a structurally plausible executable program, while a subsequent local search stage refines the program to improve geometric fidelity.

\subsubsection{\texorpdfstring{$\Pi$-StepCAD: State-Conditioned CAD Policy}{Pi-StepCAD: State-Conditioned CAD Policy}}

\textbf{Sequential generation.}
Direct optimization in program space is combinatorial and highly non-convex. We instead learn an autoregressive policy
\[
\pi_\theta(a_t \mid Q_{t-1}, Q_m, a_{<t}),
\]
where $Q_m$ is a point cloud sampled from the target mesh, $Q_{t-1}$ is sampled from the intermediate geometry, and $a_{<t}$ denotes previously generated construction actions. This induces a distribution over executable CAD programs:
\[
\pi_\theta(p)=\prod_{t=1}^T \pi_\theta(a_t \mid Q_{t-1},Q_m,a_{<t}).
\]

\textbf{Geometry conditioning.}
Target and intermediate geometries are encoded using a shared Point-MAE encoder \cite{pang2023masked}:
\[
z_m=f_\phi(Q_m), \qquad z_{t-1}=f_\phi(Q_{t-1}),
\]
and injected as conditioning tokens into the autoregressive model.

\textbf{Policy architecture.}
We instantiate $\pi_\theta$ using Qwen2.5-Coder-1.5B \cite{hui2024qwen2} conditioned on geometry embeddings and prior program tokens. The model predicts executable construction actions directly in token space.

\textbf{Training.}
The policy is trained with teacher forcing over dense state-action trajectories extracted from ARCADE-1.5M:
\[
\mathcal{L}(\theta)=
-\mathbb{E}\left[
\log \pi_\theta(a_t \mid Q_{t-1},Q_m,a_{<t})
\right].
\]
Supervision at every intermediate construction step improves long-horizon stability and geometric consistency.

\textbf{Closed-loop inference.}
At inference, generation alternates between policy prediction and CAD execution. Starting from an empty state, the model performs inference-time sampling to predict the next construction action, executes the updated CadQuery program, samples the resulting geometry, and reconditions on the produced intermediate shape before continuing generation. Generation terminates once reconstruction quality plateaus, an end token is emitted, or a cap of $T_{\max}=12$ construction actions is reached (Details in Appendix \ref{app:policy_details}). This produces a seed program $\hat p$ that captures the global structure of the target shape and initializes the refinement stage.

\subsubsection{Geometry-Aware Program Optimization}

Given a seed CAD program $\hat{p} \sim \pi_\theta$, we perform local search in program space:
\[
p^* = \arg\max_{p \in \mathcal{N}(\hat{p})} \ \mathrm{IoU}(S(p), m),
\]
where $S(p)$ denotes the executed geometry and $\mathcal{N}(\hat{p})$ is defined by local program edits.

Direct optimization over all edits is exponential in the program length $T$. We therefore perform a geometry-aware beam search over operations in execution order. At step $t$, each candidate program in the beam is expanded using a small set of local edits applied to operation $a_t$, executed in CadQuery, and ranked by IoU against the input mesh: $\mathcal{B}_t =
\mathrm{TopK}_K
\left\{
p' \in \mathrm{Expand}(p,t)
:\ p \in \mathcal{B}_{t-1}
\right\}$.

We use beam width $K=2$ and four local edits per expansion, giving a number of edit expansions linear in $T$ for fixed $K$, edit count, and per-edit budget. The final refined program is selected as:
\[
p^* = \arg\max_{p \in \mathcal{B}_T} \ \mathrm{IoU}(S(p), m).
\]

The local edits are: 
\begin{itemize}
    \item \textbf{Continuous refinement:} optimize continuous operation parameters (e.g., depths, radii, sketch dimensions). 
    \item \textbf{Sketch modification:} replace operation sketches using geometry extracted from the input mesh. 
    \item \textbf{Complement cut:} remove overshoot regions using automatically generated cut-extrudes. 
    \item \textbf{Operation skipping:} remove redundant or harmful operations.
\end{itemize}

\textbf{Implementation Details.} All algorithmic and training details are provided in appendices \ref{app:policy_details} and \ref{app:treesearch}.

\section{Experiments, Results, and Discussion}

We evaluate \textbf{StepCAD} on multiple mesh-to-CAD reconstruction benchmarks, comparing against state-of-the-art methods and analyzing the contributions of each component of our approach. We focus on geometric fidelity, validity of generated programs, and robustness to complex constructions.

\textbf{Benchmark.}
We evaluate on \textbf{CADBench} \cite{doris2026cadbench}, a comprehensive benchmark for CAD program generation that aggregates five representative datasets spanning diverse geometric domains and construction complexity. Specifically, CADBench includes: \textbf{DeepCAD} \cite{wu2021deepcad} and \textbf{Fusion 360} \cite{willis2021fusion}, \textbf{ABC} \cite{koch2019abc} for large-scale CAD shapes with varying difficulty (Easy, Medium, Hard, based on number of STEP faces \cite{contero2023quantitative}), \textbf{MCB} \cite{kim2020large} for mechanical components, and \textbf{Objaverse} \cite{deitke2023objaverse} for organic shapes. Each benchmark contains 3{,}000 samples, except ABC which provides 2{,}000 samples per complexity split. This suite enables robust evaluation across both structured mechanical designs and free-form geometries, providing a holistic assessment of CAD generation performance.

\textbf{Metrics.} We report \textbf{Intersection-over-Union (IoU)} measuring volumetric overlap, \textbf{Chamfer Distance (CD)} measuring surface discrepancy (lower is better), and \textbf{Valid Shape Rate (VSR)}, the percentage of generated programs that execute successfully and yield valid geometry. (Appendix \ref{app:metrics})

\textbf{Baselines.}
We compare against CAD-Recode \cite{rukhovich2025cad}, CADEvolve \cite{elistratov2026CADEvolve}, and Cadrille \cite{kolodiazhnyi2025Cadrille} which are representative learning-based methods for CAD generation from geometry, as well as CADReasoner \cite{kabisov2026cadreasoner}, an iterative CAD refinement for mesh to CAD generation. The main benchmark table evaluates publicly released checkpoints. Because these models use different training corpora, we additionally train a CAD-Recode-style single-shot Qwen2.5 baseline on ARCADE-1.5M (Table~\ref{tab:stepcad-ablations}). This baseline predicts a complete program from the target point cloud without intermediate-state conditioning or refinement.

\textbf{Quantitative Results.} Table~\ref{tab:mean_metrics} reports IoU and Chamfer Distance on CADBench across structured, mechanical, and organic domains. StepCAD achieves SOTA performance on all benchmarks, with the largest gains on complex shapes.

\begin{table*}[htb]
\centering
\small
\setlength{\tabcolsep}{3.2pt}
\renewcommand{\arraystretch}{1.12}
\caption{\textbf{Mean mesh-to-CAD generation metrics.}
IoU and VSR are higher better; CD is lower better.
Best overall in bold; second best underlined.}
\label{tab:mean_metrics}

\resizebox{\textwidth}{!}{%
\begin{tabular}{lccc ccc ccc ccc ccc ccc ccc}
\toprule

& \multicolumn{3}{c}{\textbf{DeepCAD}}
& \multicolumn{3}{c}{\textbf{Fusion 360}}
& \multicolumn{3}{c}{\textbf{ABC Easy}}
& \multicolumn{3}{c}{\textbf{ABC Medium}}
& \multicolumn{3}{c}{\textbf{ABC Hard}}
& \multicolumn{3}{c}{\textbf{Mechanical}}
& \multicolumn{3}{c}{\textbf{Organic}} \\

\cmidrule(lr){2-4}
\cmidrule(lr){5-7}
\cmidrule(lr){8-10}
\cmidrule(lr){11-13}
\cmidrule(lr){14-16}
\cmidrule(lr){17-19}
\cmidrule(lr){20-22}

\textbf{Model}
& IoU $\uparrow$ & CD $\downarrow$ & VSR $\uparrow$
& IoU $\uparrow$ & CD $\downarrow$ & VSR $\uparrow$
& IoU $\uparrow$ & CD $\downarrow$ & VSR $\uparrow$
& IoU $\uparrow$ & CD $\downarrow$ & VSR $\uparrow$
& IoU $\uparrow$ & CD $\downarrow$ & VSR $\uparrow$
& IoU $\uparrow$ & CD $\downarrow$ & VSR $\uparrow$
& IoU $\uparrow$ & CD $\downarrow$ & VSR $\uparrow$ \\

\midrule

CAD-Recode
& 0.79 & 0.0207 & 96.22
& 0.65 & 0.0268 & 92.63
& 0.76 & 0.0237 & 95.02
& 0.37 & 0.0349 & 88.35
& 0.35 & 0.0328 & 87.57
& 0.66 & 0.0297 & 94.01
& 0.40 & 0.0570 & 87.84 \\

Cadrille
& 0.78 & 0.0211 & 96.97
& 0.66 & 0.0268 & 94.87
& 0.75 & 0.0234 & 96.58
& 0.40 & 0.0359 & 92.30
& 0.37 & 0.0337 & 92.18
& 0.65 & 0.0295 & 96.80
& 0.42 & 0.0549 & 91.03 \\

CADEvolve
& 0.74 & 0.0389 & 99.70
& 0.63 & 0.0466 & 97.60
& 0.72 & 0.0413 & 98.88
& 0.39 & 0.0638 & 95.82
& 0.39 & 0.0617 & 95.27
& 0.64 & 0.0486 & 96.90
& 0.40 & 0.0890 & 96.91 \\

CADReasoner
& \underline{0.86} & \underline{0.0197} & 99.47
& 0.71 & 0.0326 & \underline{98.33}
& \underline{0.83} & 0.0242 & 99.60
& 0.43 & 0.0419 & 97.10
& 0.31 & 0.0627 & 92.80
& 0.67 & 0.0413 & 98.80
& 0.31 & 0.0968 & 96.60 \\

\midrule

$\Pi$-StepCAD \textbf{(Ours)}
& 0.82 & 0.0205 & \textbf{100.00}
& \underline{0.75} & \underline{0.0240} & \textbf{99.97}
& 0.77 & \underline{0.0227} & \textbf{100.00}
& \underline{0.53} & \underline{0.0320} & \textbf{100.00}
& \underline{0.52} & \underline{0.0283} & \textbf{99.95}
& \underline{0.78} & \underline{0.0259} & \textbf{100.00}
& \underline{0.53} & \underline{0.0446} & \textbf{100.00} \\

StepCAD \textbf{(Ours)}
& \textbf{0.89} & \textbf{0.0191} & \textbf{100.00}
& \textbf{0.83} & \textbf{0.0228} & \textbf{99.97}
& \textbf{0.89} & \textbf{0.0200} & \textbf{100.00}
& \textbf{0.74} & \textbf{0.0302} & \textbf{100.00}
& \textbf{0.73} & \textbf{0.0263} & \textbf{99.95}
& \textbf{0.89} & \textbf{0.0213} & \textbf{100.00}
& \textbf{0.71} & \textbf{0.0420} & \textbf{100.00} \\

\midrule

\makecell{Improvement \\ vs. Best Prior} (\%)
& 3.5 & 3.0 & 0.3
& 16.9 & 14.9 & 1.7
& 7.2 & 14.5 & 0.4
& 72.1 & 13.5 & 3.0
& 87.2 & 19.8 & 4.9
& 32.8 & 27.8 & 1.2
& 69.0 & 23.5 & 3.2 \\

\bottomrule
\end{tabular}
}
\end{table*}

\textbf{State-of-the-art mesh-to-CAD reconstruction.}
StepCAD consistently outperforms prior methods across all datasets. On \textbf{ABC Hard}, IoU improves from $0.31$--$0.39$ to \textbf{0.73}, an \textbf{87.2\%} relative gain over the strongest baseline. On \textbf{ABC Medium}, IoU increases from $0.43$ to \textbf{0.74} (\textbf{72.1\%} gain). On the \textbf{Mechanical} and \textbf{Organic} subsets, StepCAD reaches \textbf{0.89} and \textbf{0.71} IoU, improving by \textbf{32.8\%} and \textbf{69.0\%} respectively. StepCAD also achieves the lowest Chamfer Distance on every benchmark, with reductions up to \textbf{28.0\%} on Mechanical, demonstrating robustness across both structured CAD models and free-form geometries.

\textbf{Benefits of state-conditioned step-by-step generation.}
Even without optimization, $\Pi$-StepCAD already surpasses prior work, especially on difficult benchmarks (e.g., $0.52$ IoU vs.\ $\sim0.38$ on ABC Hard, $\sim37\%$ improvement). Unlike one-shot autoregressive approaches, $\Pi$-StepCAD conditions each prediction on the intermediate geometry produced so far, enabling progressively more accurate structural and parametric decisions throughout construction. We note that CADReasoner performs slightly better than $\Pi$-StepCAD on the easiest benchmarks (DeepCAD and ABC Easy), likely due to specialization to simpler in-distribution samples. However, its performance degrades substantially on harder subsets, whereas $\Pi$-StepCAD remains robust. This suggests that explicit intermediate-geometry conditioning generalizes better to complex shapes than iterative learned refinement alone.

\textbf{Generative optimization is critical.}
The full StepCAD system substantially improves over the policy alone (e.g., $0.52 \rightarrow 0.73$ IoU on ABC Hard). The geometry-aware tree search refines program structure and numerical parameters (e.g., extrusion depths and sketch dimensions), producing an average IoU gain of $0.14$ ($\sim20\%$ relative). These results show that combining learned proposals with targeted program-space search is essential for high-fidelity CAD reconstruction.

\textbf{Surface accuracy.}
Chamfer Distance follows the same trend. StepCAD achieves the lowest error on all datasets (e.g., $0.0263$ vs.\ $0.0283$ on ABC Hard, and $0.0420$ vs.\ $0.0446$ on Organic), indicating improvements not only in volumetric overlap but also in fine geometric detail.

\textbf{Disentangling training data, policy, and optimization.}
Table~\ref{tab:stepcad-ablations} adds a single-shot Qwen2.5 model fine-tuned on ARCADE-1.5M using a CAD-Recode-style point-cloud-to-program setup. It reaches $0.644$ IoU and $0.041$ CD without intermediate-geometry conditioning or search, compared with $0.568$ IoU for the released CADEvolve checkpoint. This demonstrates that the corpus supports a conventional generator, although differences in model and training setup prevent attributing that cross-model gap solely to data. Within StepCAD, removing previous-geometry tokens reduces IoU from $0.682$ to $0.632$. The full policy exceeds the single-shot baseline by $0.038$ IoU, and program-space search further improves the policy from $0.682$ to $0.816$ ($+0.134$). Together, the experiments distinguish the benefits of an operation-rich corpus, intermediate-state conditioning, and explicit refinement without treating the released-checkpoint comparison as a controlled data-only ablation.

\begin{table}[H]
\centering
\small
\setlength{\tabcolsep}{6pt}
\renewcommand{\arraystretch}{1.1}
\caption{\textbf{Training data, state conditioning, and search.} All variants except the released CADEvolve checkpoint use ARCADE-1.5M. The single-shot baseline predicts a complete program without intermediate geometry.}
\label{tab:stepcad-ablations}
\begin{tabular}{lccc}
\toprule
\textbf{Variant} & \textbf{CD $\downarrow$} & \textbf{IoU $\uparrow$} & \textbf{VSR (\%)}$\uparrow$ \\
\midrule
CADEvolve (public checkpoint) & 0.056 & 0.568 & 97 \\
Single-shot Qwen2.5 on ARCADE-1.5M & 0.041 & 0.644 & 100 \\
$\Pi$-StepCAD w/o previous-geometry tokens   & 0.045 & 0.632 & \textbf{100} \\
\textbf{$\Pi$-StepCAD}                       & \textbf{0.028} & \textbf{0.682} & \textbf{100} \\
\textbf{StepCAD}                       & \textbf{0.025} & \textbf{0.816} & \textbf{100} \\
\bottomrule
\end{tabular}
\end{table}

\textbf{Additional evaluation: fidelity, search transfer, and compactness.}
Appendix~\ref{app:rebuttal-evaluation} complements the main benchmark results with high-IoU success rates, entity-level geometric accuracy, search transferability, and program compactness. StepCAD achieves IoU $\geq0.95$ on $39.9\%$ of CADBench samples and IoU $\geq0.99$ on $15.3\%$, compared with $27.2\%$ and $7.4\%$ for the strongest baseline, CADReasoner (Table~\ref{tab:high-iou}). It also improves vertex, edge, and face accuracy from $15.7$, $9.7$, and $6.4$ for the strongest baseline, CADEvolve, to $20.8$, $12.6$, and $8.1$, respectively (Table~\ref{tab:entity-accuracy}). Applying our search to CAD-Recode programs on 1,200 samples improves mean IoU from $0.597$ to $0.745$ without retraining (Table~\ref{tab:search-transfer}). Refinement increases mean program size from 16.2 to 31.7 counted operations as reconstruction fidelity improves (Table~\ref{tab:program-size}); on DeepCAD, the refined programs have an overall recovered/reference length ratio of $0.91$ at $0.89$ IoU (Table~\ref{tab:conciseness}). These evaluations distinguish geometric fidelity and compactness from recovery of a unique parameterization or design intent.

\begin{figure}[H]
    \centering
    \includegraphics[width=1\linewidth]{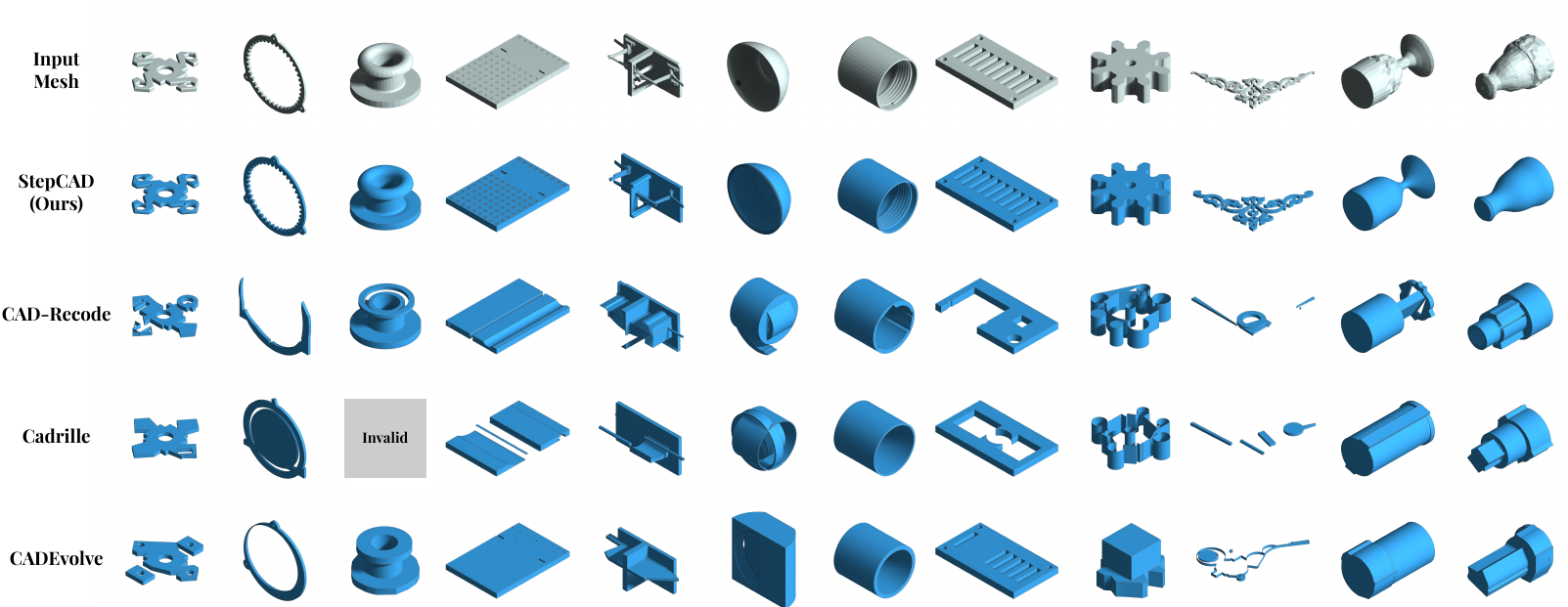}
    \caption{
    \textbf{Qualitative mesh-to-CAD reconstruction results.} Rows show input mesh, StepCAD, CAD-Recode, Cadrille, and CADEvolve; ‘Invalid’ denotes a failed/non-executable reconstruction.
    }
    \label{fig:qualitative}
\end{figure}

\textbf{Qualitative results.}
Appendix~\ref{app:trajectories} provides construction trajectories (Fig.~\ref{fig:construction-trajectories}), competing search edits and finishing-feature examples (Fig.~\ref{fig:search-edit-comparison}), failure cases (Fig.~\ref{fig:failure-cases}), evaluation inputs (Fig.~\ref{fig:evaluation-input-gallery}), and expanded policy-versus-refinement comparisons (Fig.~\ref{fig:qualitative-expanded}). Figure~\ref{fig:qualitative} shows reconstructions across datasets. StepCAD produces CAD programs that closely match target geometry while preserving both global structure and fine details.

\textbf{Impact of ARCADE-1.5M and operation diversity.}
A key factor behind these gains is the broader operation coverage of \textbf{ARCADE-1.5M}, including fillets, chamfers, lofts, and revolves. Earlier sketch-extrude corpora offer narrower coverage, while recent corpora such as CADEvolve and Zero-to-CAD also contain finishing and other solid operations. Their presence in training data should be distinguished from their frequency in generated programs. As a result, StepCAD generates more diverse operations. On the \textbf{Organic} subset, lofts appear in $30.4\%$ of generated programs versus $3.0\%$ for CADEvolve, while revolves appear in $23.2\%$ versus $0.8\%$. Prior methods rely overwhelmingly on extrusions. (Appendix~\ref{app:generated-seqs})

This diversity is critical for reconstructing geometries that cannot be represented well with sketch-extrude primitives alone. For example, in the last column of Fig.~\ref{fig:qualitative}, StepCAD uses loft operations to recover smooth non-planar transitions, achieving higher fidelity (\textbf{IoU: $0.83$}, \textbf{CD: $0.066$}) than CADEvolve (\textbf{IoU: $0.66$}, \textbf{CD: $0.108$}) and Cadrille (\textbf{IoU: $0.67$}, \textbf{CD: $0.101$}), which primarily rely on extrusion-based constructions. These results demonstrate that operation diversity is essential for faithful reconstruction of complex geometry.

\textbf{Applications and one-to-many CAD generation.}
Beyond reconstruction, StepCAD enables several downstream applications enabled by its state-conditioned generative formulation. As shown in Appendix~\ref{app:applications} and Fig.~\ref{fig:variants}, $\Pi$-StepCAD can generate diverse editable CAD variants from a single reference mesh while preserving the overall structure, enabling rapid design exploration and parametric shape variation. StepCAD can also be integrated into image-to-CAD pipelines (Appendix~\ref{app:applications}, Fig.~\ref{fig:imagetocad}) by reconstructing parametric CAD programs from meshes generated by image-to-3D models such as Hunyuan3D~\cite{zhao2025hunyuan3d}. Finally, Appendix~\ref{app:one-to-many} and Fig.~\ref{fig:one-to-many} highlight the inherently one-to-many nature of mesh-to-CAD reconstruction: the policy can generate multiple valid CAD programs with different operation sequences that produce nearly identical geometry. This suggests that StepCAD learns a distribution over plausible CAD construction strategies rather than memorizing a single deterministic reconstruction process.

Overall, these results demonstrate the effectiveness of combining state-conditioned generative policies with geometry-aware optimization for robust and accurate CAD program synthesis.

\textbf{Additional ablations.}
Additional ablations are provided in the appendix, including the effects of Qwen model scale and point-cloud encoder choice (Appendix~\ref{app:ab-model-encoder}, Table~\ref{tab:design_sweep}); policy inputs, Point-MAE fine-tuning, and point sampling strategy (Appendix~\ref{app:ab-stepcad}, Table~\ref{tab:ablations}); and the contribution of each local edit within the tree-search optimization procedure (Appendix~\ref{app:ab-search}, Table~\ref{tab:phase-ablation}).

\section{Limitations and Future Work}
\label{sec:limitations}

\textbf{Computation and search.} Refinement takes approximately one minute per sample and relies on a fixed set of local edits. Its bounded parameter neighborhoods and finite beam may miss a useful correction or fail to recover from a poor seed. The 12-action policy cap limits generation even though one action may contain multiple counted operations; training sequences longer than this cap do not establish equivalent inference-time capacity. \textbf{Geometric versus semantic fidelity.} IoU and CD measure shape agreement, not recovery of original dimensions, design intent, constraints, or feature semantics. In particular, removing a misparameterized but meaningful fillet may improve IoU if a better radius is not found. \textbf{Selection and editability.} Coarse edge and directional face selectors cannot address arbitrary individual entities. Programs are executable and their code parameters can be edited, but we do not establish constraint preservation or user editing success. Program compactness is only one aspect of editability. \textbf{Data and generalization.} Synthetic training introduces a domain gap, and alternative valid histories complicate comparison with a reference B-rep. Future work should combine aligned point-cloud/B-rep representations, entity-specific operators, feature-aware objectives, constraint modeling, and user studies, while improving search efficiency. \textbf{Societal impact.} Reconstruction can facilitate replication of proprietary designs and change designer workflows, motivating attribution and human oversight.

\section{Conclusion}
\label{sec:conclusion}

We presented \textbf{StepCAD}, a generative optimization approach for mesh-to-CAD reconstruction that combines a state-conditioned CAD policy with geometry-aware program refinement. By reformulating CAD generation as optimization in program space, our approach overcomes the limitations of one-shot sequence prediction and achieves SOTA performance across multiple benchmarks. We further introduced \textbf{ARCADE-1.5M}, a large-scale dataset of CAD programs with diverse operations and long-horizon construction sequences, enabling effective training of state-conditioned models. Our results demonstrate that combining learned generation with explicit geometric optimization leads to substantial improvements in fidelity, robustness, and generalization. More broadly, this work highlights the importance of integrating generative models with search and optimization for structured program synthesis tasks. We believe this paradigm can be extended beyond CAD reconstruction to applications such as image-to-CAD generation, interactive design, and automated engineering workflows.

\newpage

\bibliographystyle{unsrt}
\bibliography{my_bib}


\newpage
\appendix
\addcontentsline{toc}{part}{Appendices}

\etocsettocstyle{\large\textbf{Table of Contents for Appendices}\par\vspace{1em}}{}

\makeatletter
\begingroup
  \let\oldlsection\l@section
  \let\oldlsubsection\l@subsection
  \let\oldlsubsubsection\l@subsubsection 
  
  \renewcommand{\l@section}[2]{\oldlsection{\small #1}{\small #2}}
  \renewcommand{\l@subsection}[2]{\oldlsubsection{\small #1}{\small #2}}
  \renewcommand{\l@subsubsection}[2]{\oldlsubsubsection{\small #1}{\small #2}} 
  
  \localtableofcontents
\endgroup
\makeatother

\vspace{2em}

\newpage

\section{Ablations}

\subsection{Which Components of $\Pi$-StepCAD Matter Most?}
\label{app:ab-stepcad}
\begin{table}[H]
\centering
\small
\setlength{\tabcolsep}{6pt}
\renewcommand{\arraystretch}{1.1}
\caption{\textbf{Ablation study of StepCAD.} CD $\downarrow$ and IoU $\uparrow$ evaluate reconstruction quality; VSR denotes valid shape rate.}
\label{tab:ablations}
\begin{tabular}{lccc}
\toprule
\textbf{Variant} & \textbf{CD $\downarrow$} & \textbf{IoU $\uparrow$} & \textbf{VSR (\%)}$\uparrow$ \\
\midrule
$\Pi$-StepCAD w/o prompt conditioning       & --    & --    & 0   \\
$\Pi$-StepCAD w/o FPS point sampling         & 0.209 & 0.105 & 100 \\
$\Pi$-StepCAD w/o input geometry tokens      & 0.144 & 0.209 & 100 \\
$\Pi$-StepCAD w/o Point-MAE fine-tuning       & 0.093 & 0.316 & 100 \\
CADEvolve (public checkpoint) & 0.056 & 0.568 & 97 \\
Single-shot Qwen2.5 on ARCADE-1.5M & 0.041 & 0.644 & 100 \\
$\Pi$-StepCAD w/o previous-geometry tokens   & 0.045 & 0.632 & 100 \\
\textbf{$\Pi$-StepCAD}                       & \textbf{0.028} & \textbf{0.682} & 100 \\
\textbf{StepCAD}                       & \textbf{0.025} & \textbf{0.816} & 100 \\
\bottomrule
\end{tabular}
\end{table}

\subsection{How Do Model Scale and Geometry Encoders Affect Performance?}
\label{app:ab-model-encoder}
\begin{table}[H]
\centering
\small
\setlength{\tabcolsep}{6pt}
\renewcommand{\arraystretch}{1.1}
\caption{\textbf{Ablation study for model size and point cloud encoder for $\Pi$-StepCAD.} Best validation loss achieved within the first 10K training steps. Lower is better.}
\label{tab:design_sweep}
\begin{tabular}{lc}
\toprule
\textbf{Variant} & \textbf{Best Val Loss at 10K steps $\downarrow$} \\
\midrule
$\Pi$-StepCAD (0.5B + Point-MAE)            & 0.3559 \\
\underline{$\Pi$-StepCAD (1.5B + Point-MAE)} & \underline{0.3433} \\
\textbf{$\Pi$-StepCAD (3B + Point-MAE)}      & \textbf{0.3419} \\
$\Pi$-StepCAD (1.5B + Utonia \cite{zhang2026utonia})              & 0.3450 \\
\bottomrule
\end{tabular}
\end{table}

Although the 3B model achieves the lowest early validation loss, its gain over the 1.5B model is marginal ($0.0014$), so we use the 1.5B Point-MAE configuration as the default due to its favorable compute-performance trade-off.

\subsection{Which Local Edits Contribute Most to Geometry-Aware Search?}
\label{app:ab-search}
\begin{table}[H]
  \centering\small
  \caption{\textbf{Local edit-level ablation of the tree search.} Mean metrics; \\
  \textbf{Win-rate}: fraction of beam steps where this phase's output is the rank-0 candidate (i.e.\ the new beam[0]). \\
  \textbf{Realized $\Delta$~IoU}: average per-sample IoU lift attributed to this phase via winner-attribution at each step; the four phase rows sum to the \emph{Full search} row by construction. \\
  \textbf{\% share}: the phase's Realized $\Delta$ divided by the Full Realized $\Delta$. \\
  \textbf{Cost}: mean wall-time per sample.}

  \label{tab:phase-ablation}
  \begin{tabular}{lcccc}
    \toprule
    Local Edit & Win-rate & Realized $\Delta$ & \% share & Cost (s) \\
    \midrule
    Sketch Modification    & 26.0\% & \textbf{+0.059} & 41.5\% & 12.5 \\
    Complement Cut         & 24.4\% & +0.043 & 30.3\% & 17.0 \\
    Continuous refinement  & 36.6\% & +0.023 & 16.2\% & 28.4 \\
    Operation skipping     & 13.0\% & +0.017 & 12.0\% & 8.1 \\
    \midrule
    Full search            & 100\%  & \textbf{+0.142} & 100\% & 66.0 \\
    \bottomrule
\end{tabular}
\end{table}

\newpage
\section{StepCAD: Applications}
\label{app:applications}
\subsection{Variants Generation}

\begin{figure}[H]
    \centering
    \includegraphics[width=\linewidth]{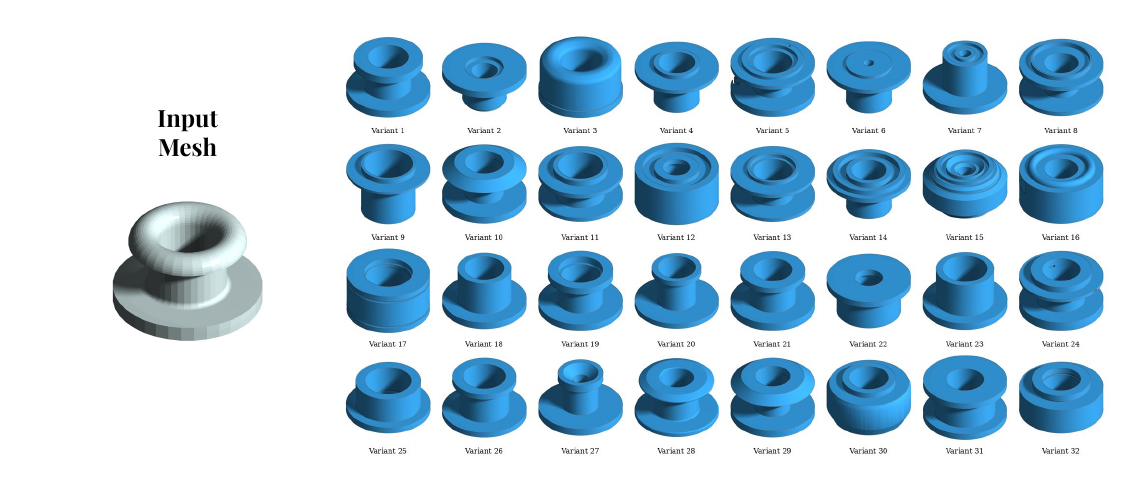}
    \caption{\textbf{Design variation generation with $\Pi$-StepCAD.} Given an input mesh, $\Pi$-StepCAD can generate diverse parametric CAD variants that preserve the overall structure while modifying geometric details and construction choices. This demonstrates a potential application of the learned policy for CAD design exploration and rapid generation of editable shape variations from a single reference geometry.}
    \label{fig:variants}
\end{figure}

\subsection{Image-to-CAD}

\begin{figure}[H]
    \centering
    \includegraphics[width=1\linewidth]{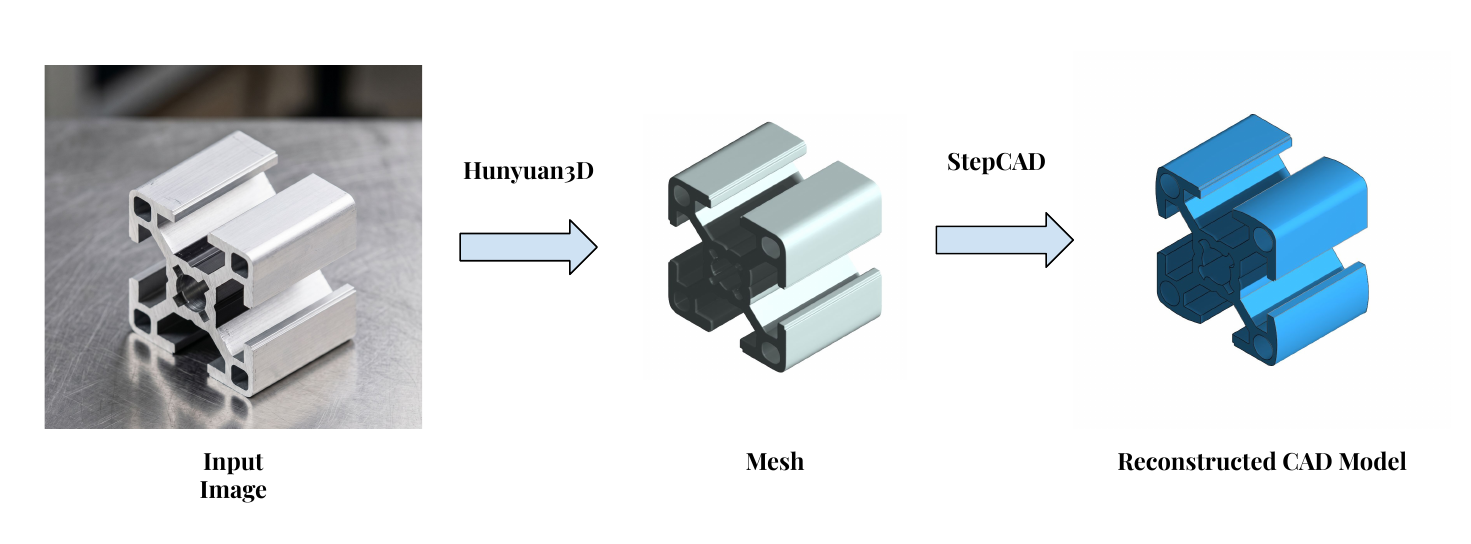}
    \caption{\textbf{Image-to-CAD reconstruction using StepCAD.} An input image is first converted into a 3D mesh using Hunyuan3D \cite{zhao2025hunyuan3d}, after which StepCAD reconstructs an editable parametric CAD program from the generated geometry. This demonstrates how StepCAD can be integrated into image-to-CAD pipelines to transform unstructured visual inputs into structured CAD representations.}
    \label{fig:imagetocad}
\end{figure}

\newpage

\section{One-to-Many Mapping: From Geometry to CAD Construction Sequence}
\label{app:one-to-many}

\begin{figure}[H]
    \centering
    \includegraphics[width=1\linewidth]{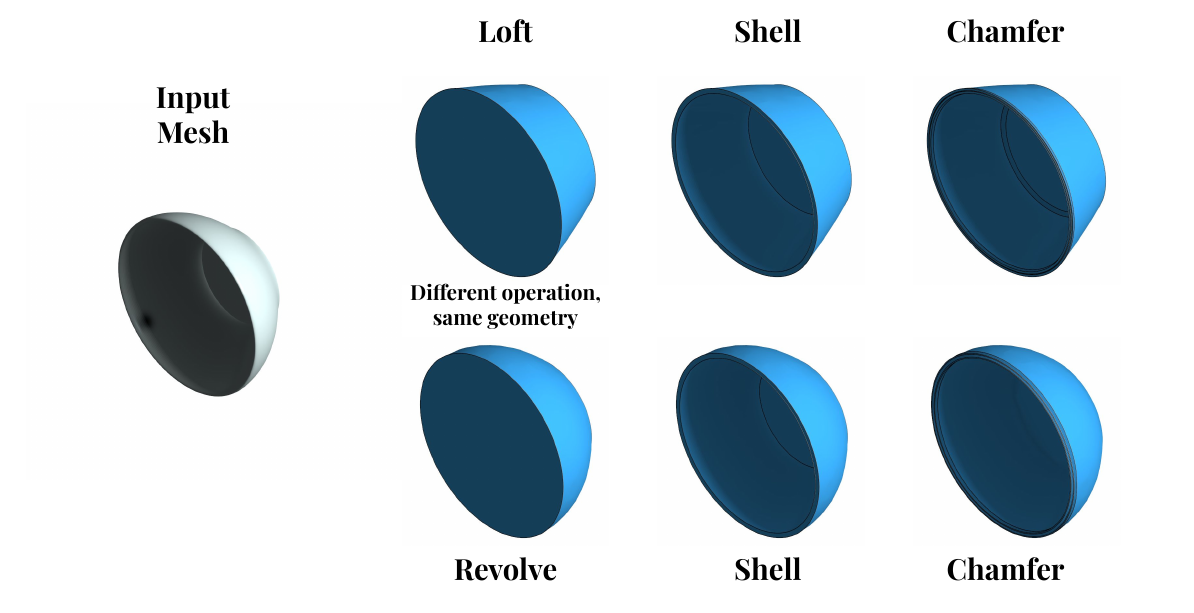}
    \caption{\textbf{One-to-many CAD program generation from a single input mesh.} Given the same target geometry, StepCAD generates multiple valid CAD programs with different operation sequences. In this example, the model reconstructs the shape using either a loft-based or revolve-based strategy, followed by shell and chamfer operations, producing nearly identical final geometry. This demonstrates that the policy learns a distribution over valid CAD construction procedures rather than a single deterministic reconstruction.}
    \label{fig:one-to-many}
\end{figure}

\section{Analysis of Generated CAD Operation Distributions}
\label{app:generated-seqs}
\begin{table}[h]
\centering
\small
\setlength{\tabcolsep}{5pt}
\renewcommand{\arraystretch}{1.1}
\caption{\textbf{Operation occurrence across generated programs (\%).} 
Values indicate the percentage of generated CAD programs that contain at least one instance of the given operation.}
\begin{tabular}{lcccccccc}
\toprule
\textbf{Model} 
& \textbf{Extrude} 
& \textbf{Revolve} 
& \textbf{Sweep} 
& \textbf{Loft} 
& \textbf{Fillet} 
& \textbf{Chamfer} 
& \textbf{Shell} 
& \textbf{Mirror} \\
\midrule
StepCAD 
& 89.35 & 16.86 & 7.75 & 11.48 & 9.60 & 6.24 & 5.11 & 2.82 \\
CADEvolve 
& 81.10 & 0.20 & 0.00 & 1.20 & 3.30 & 1.80 & 0.20 & 0.00 \\
Cadrille 
& 100.00 & 0.00 & 0.00 & 0.00 & 0.00 & 0.00 & 0.00 & 0.00 \\
CAD-Recode 
& 100.00 & 0.00 & 0.00 & 0.00 & 0.00 & 0.00 & 0.00 & 0.00 \\
\bottomrule
\end{tabular}
\end{table}

\begin{table}[H]
\centering
\small
\setlength{\tabcolsep}{4pt}
\renewcommand{\arraystretch}{1.1}
\caption{\textbf{Operation occurrence across generated programs (\%).} 
Values indicate the percentage of generated CAD programs that contain at least one instance of the given operation. }
\label{tab:operation_occurrence_benchmarks}
\begin{tabular}{llcccccccc}
\toprule
\textbf{Benchmark} & \textbf{Model}
& \textbf{Extrude} 
& \textbf{Revolve} 
& \textbf{Sweep} 
& \textbf{Loft} 
& \textbf{Fillet} 
& \textbf{Chamfer} 
& \textbf{Shell} 
& \textbf{Mirror} \\
\midrule

\multirow{2}{*}{DeepCAD}
& StepCAD 
& 94.5 & 11.3 & 4.4 & 4.9 & 4.1 & 1.7 & 2.9 & 2.2 \\
& CADEvolve 
& 80.9 & 0.0 & 0.0 & 0.0 & 0.1 & 0.0 & 0.4 & 0.0 \\
\midrule

\multirow{2}{*}{Fusion360}
& StepCAD 
& 86.7 & 19.9 & 10.2 & 12.5 & 10.0 & 4.9 & 3.3 & 3.1 \\
& CADEvolve 
& 81.7 & 0.0 & 0.0 & 0.8 & 3.1 & 1.0 & 0.1 & 0.0 \\
\midrule

\multirow{2}{*}{ABC}
& StepCAD 
& 92.0 & 14.4 & 6.1 & 8.4 & 9.9 & 5.5 & 7.4 & 2.7 \\
& CADEvolve 
& 82.9 & 0.0 & 0.0 & 0.6 & 4.8 & 1.3 & 0.2 & 0.0 \\
\midrule

\multirow{2}{*}{Mechanical}
& StepCAD 
& 75.7 & 33.0 & 6.1 & 32.2 & 23.3 & 17.6 & 5.3 & 2.3 \\
& CADEvolve 
& 70.2 & 0.0 & 0.0 & 2.5 & 5.0 & 6.8 & 0.0 & 0.0 \\
\midrule

\multirow{2}{*}{Organic}
& StepCAD 
& 79.6 & 23.2 & 21.8 & 30.4 & 12.9 & 20.4 & 6.5 & 4.8 \\
& CADEvolve 
& 88.1 & 0.8 & 0.1 & 3.0 & 1.9 & 0.6 & 0.2 & 0.0 \\
\bottomrule
\end{tabular}
\end{table}

\clearpage
\section{Construction Trajectories, Search Behavior, and Qualitative Evaluation}
\label{app:trajectories}

\begin{figure}[H]
\centering
\includegraphics[width=0.8\linewidth,keepaspectratio]{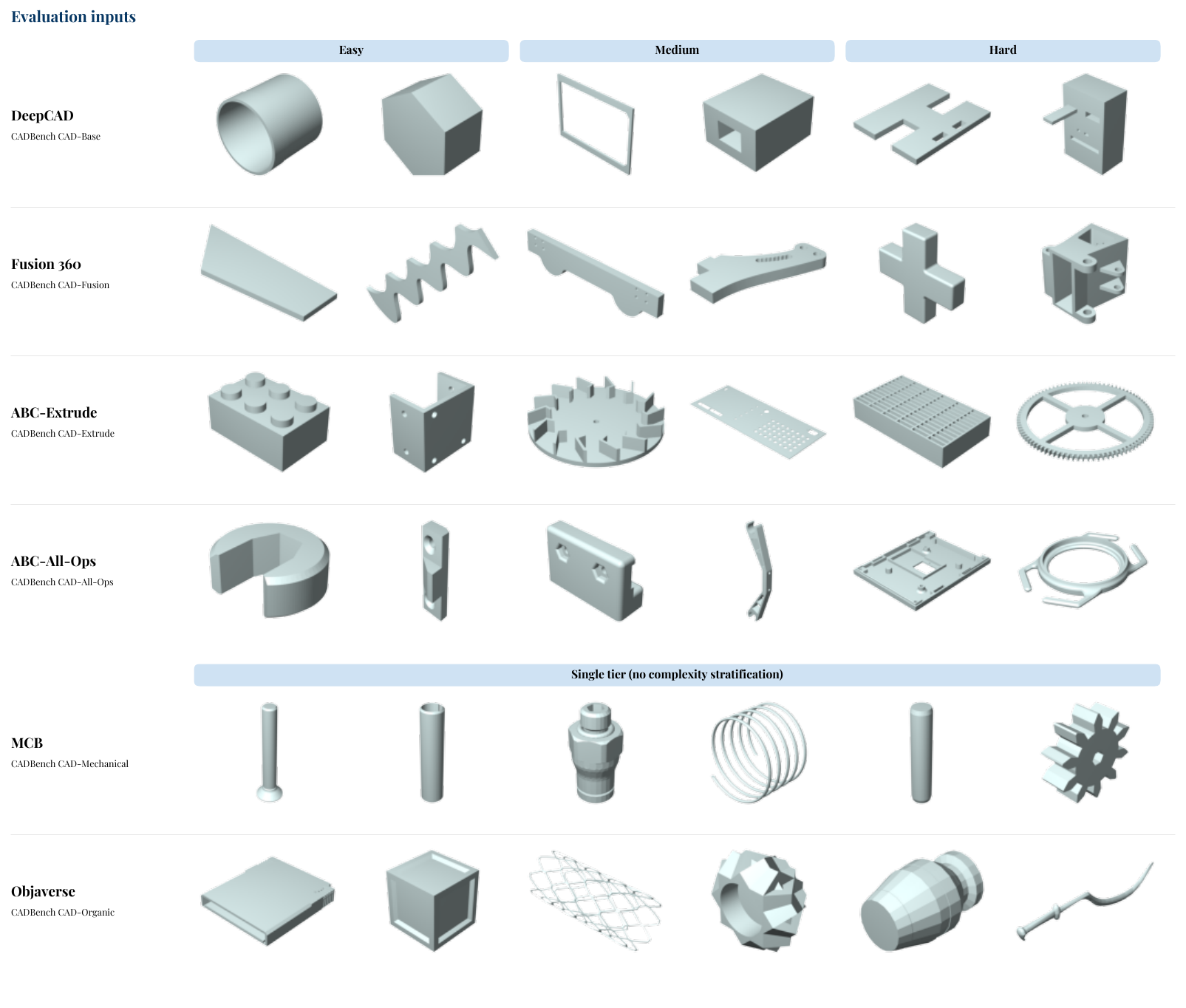}
\caption{\textbf{Evaluation-input gallery.} Example meshes from DeepCAD, Fusion 360, ABC-Extrude, ABC-All-Ops, MCB, and Objaverse, using the category labels supplied with the gallery.}
\label{fig:evaluation-input-gallery}
\end{figure}

\begin{figure}[H]
\centering
\includegraphics[width=0.9\linewidth]{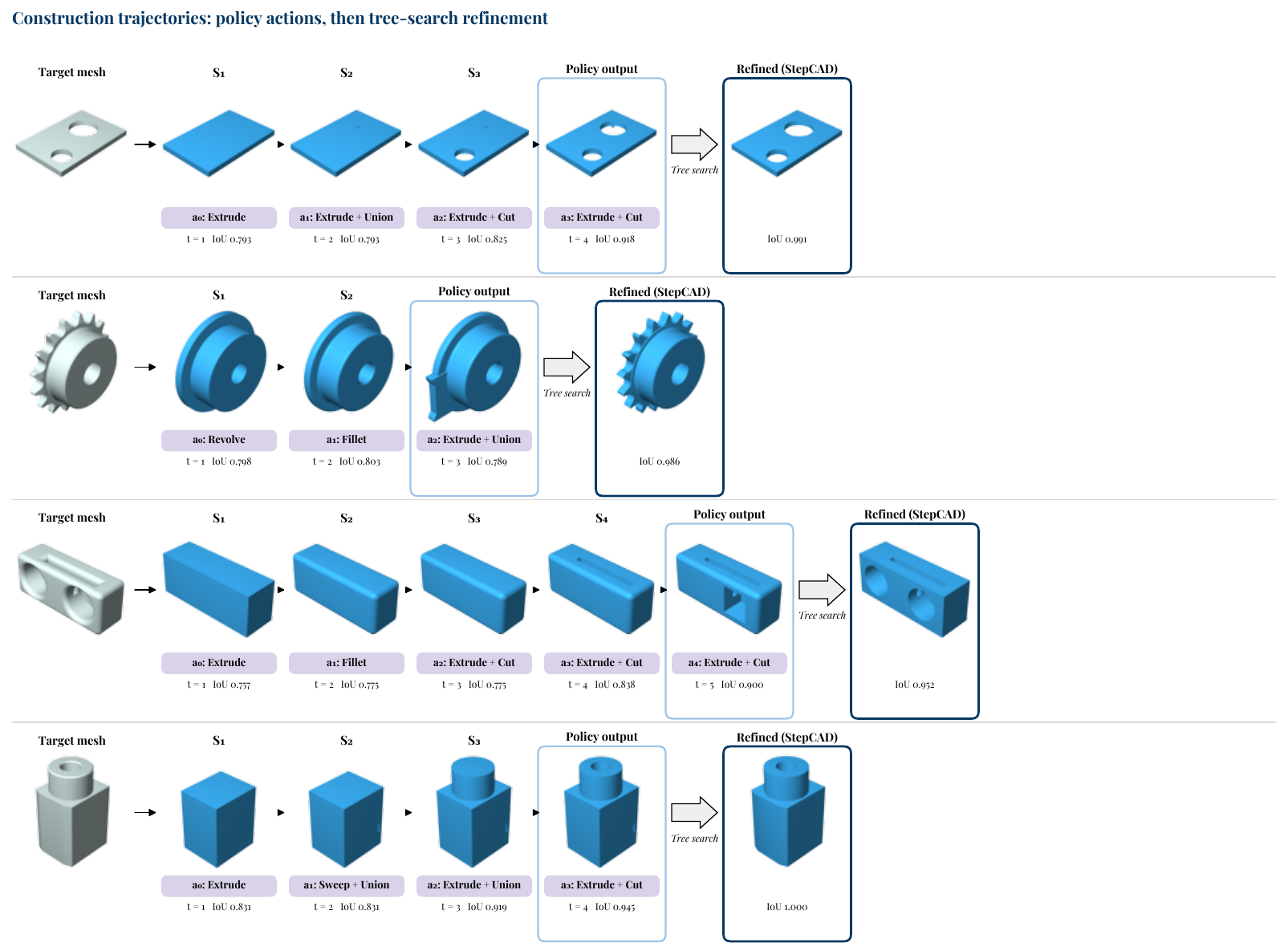}
\caption{\textbf{Construction trajectories followed by refinement.} Four examples show the target mesh, successive intermediate solids, the policy output, and the refined StepCAD reconstruction.}
\label{fig:construction-trajectories}
\end{figure}


\begin{figure}[H]
\centering
\includegraphics[width=0.7\linewidth]{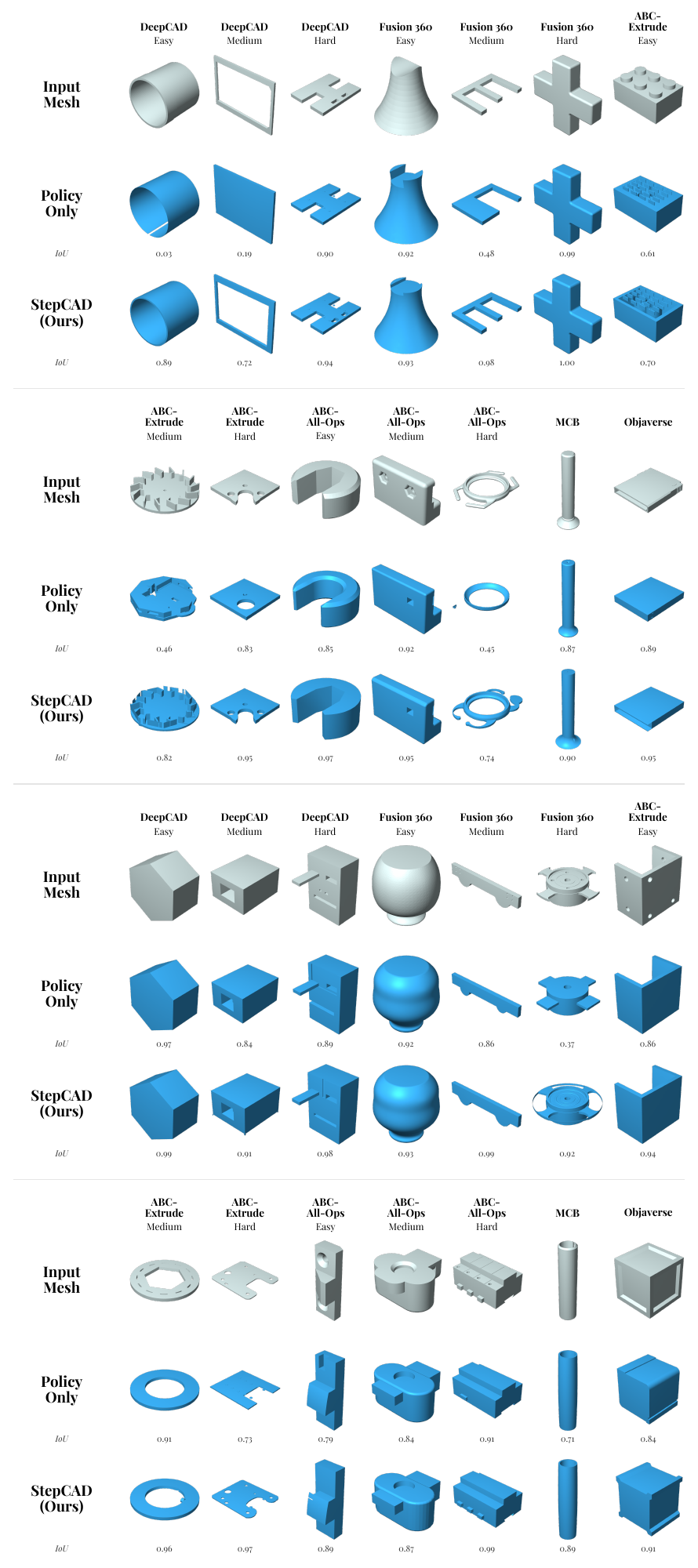}
\caption{Expanded qualitative reconstructions}
\label{fig:qualitative-expanded}
\end{figure}

\newpage

\begin{figure}[H]
\centering
\includegraphics[width=\linewidth,height=0.78\textheight,keepaspectratio]{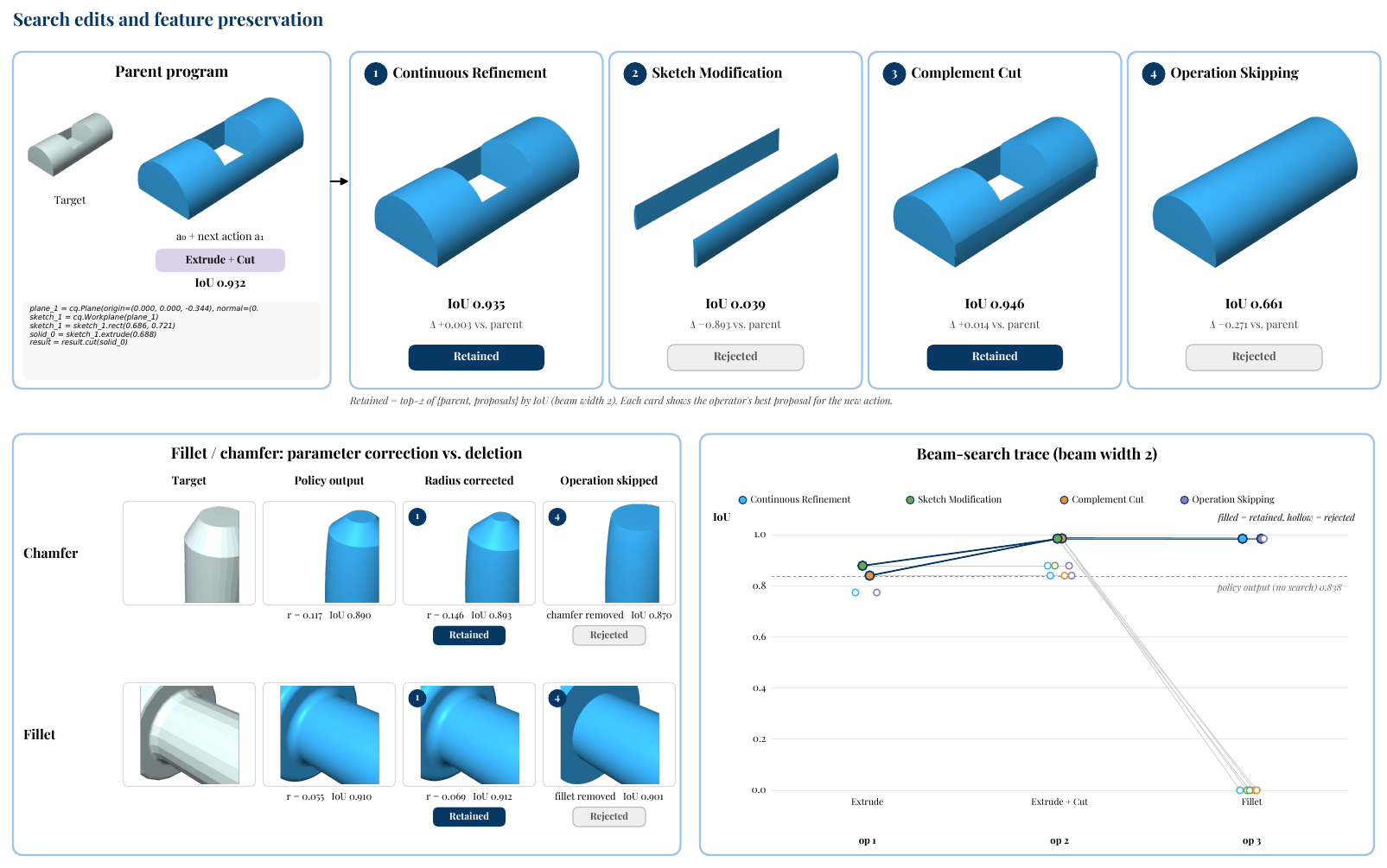}
\caption{\textbf{Competing search edits and finishing-feature examples.} The upper panels compare four edit proposals from a common parent, with IoU and retained/rejected status. The lower-left panels compare parameter correction with removal of a chamfer or fillet; correction yields higher IoU in these examples. The lower-right panel shows a beam-search trace with retained and rejected candidates.}
\label{fig:search-edit-comparison}
\end{figure}

\begin{figure}[H]
\centering
\includegraphics[width=\linewidth,height=0.78\textheight,keepaspectratio]{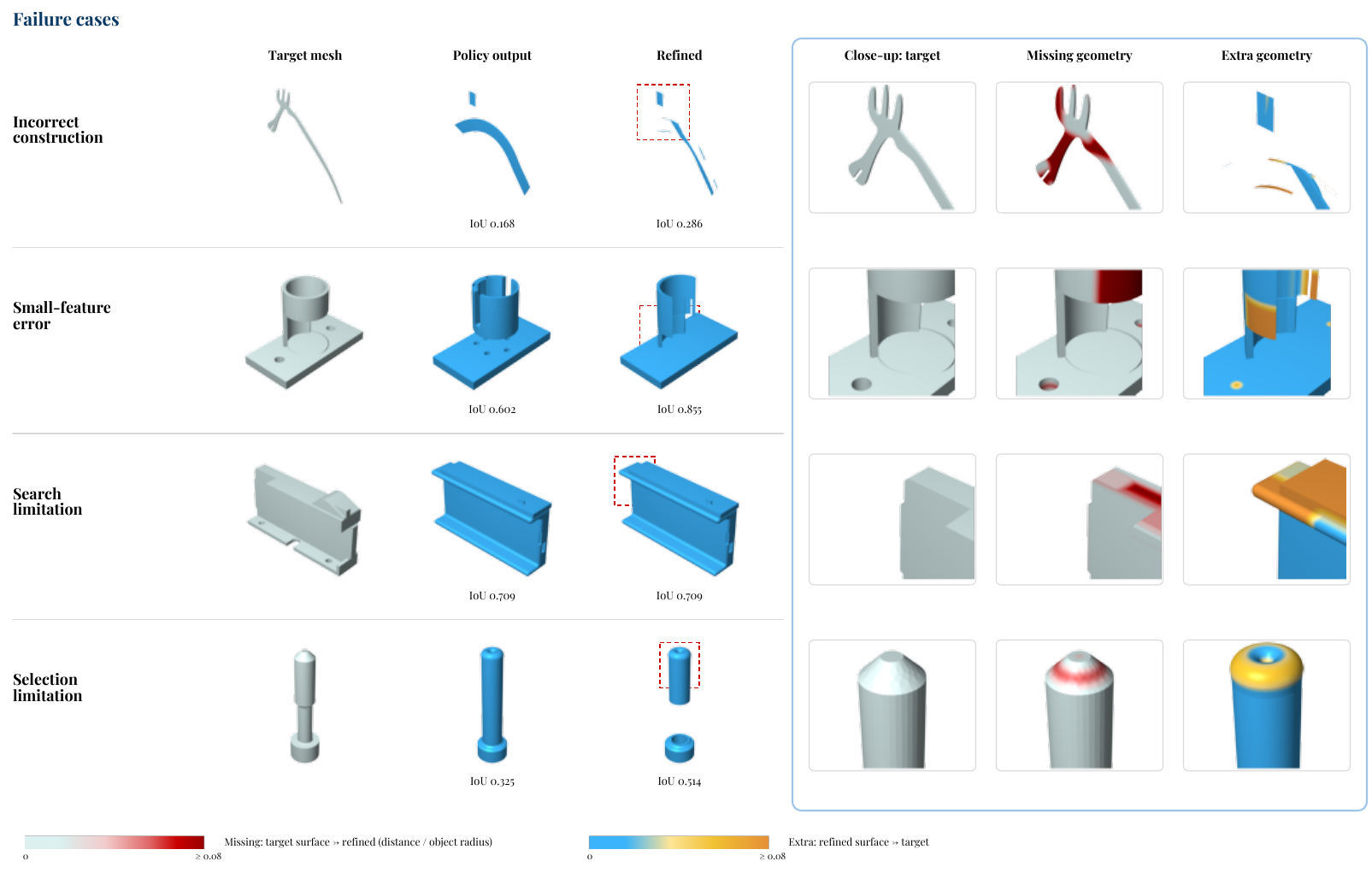}
\caption{\textbf{Residual errors and failure cases.} Target meshes, policy outputs, and refined reconstructions are shown with IoU.}
\label{fig:failure-cases}
\end{figure}

\section{Metrics}
\label{app:metrics}
We report the following metrics:
\begin{itemize}[leftmargin=1.2em]
    \item \textbf{Intersection-over-Union (IoU):} volumetric overlap between predicted and target shapes.
    \[
\text{IoU} = \frac{V(\hat{S} \cap S)}{V(\hat{S} \cup S)}
\]
    \item \textbf{Chamfer Distance (CD):} surface distance between point clouds (lower is better).
    \[
\text{CD}(\hat{P}, P) = \frac{1}{2}\left(\frac{1}{|\hat{P}|} \sum_{x \in \hat{P}} \min_{y \in P} \|x - y\|_2^2
+ \frac{1}{|P|} \sum_{y \in P} \min_{x \in \hat{P}} \|y - x\|_2^2\right)
\]
    \item \textbf{Valid Shape Rate (VSR):} percentage of generated programs that execute successfully and produce valid geometry.
\end{itemize}

\newpage

\section{Dataset Statistics}
\label{app:dataset-stats}

\begin{figure}[H]
    \centering
    \includegraphics[width=1\linewidth]{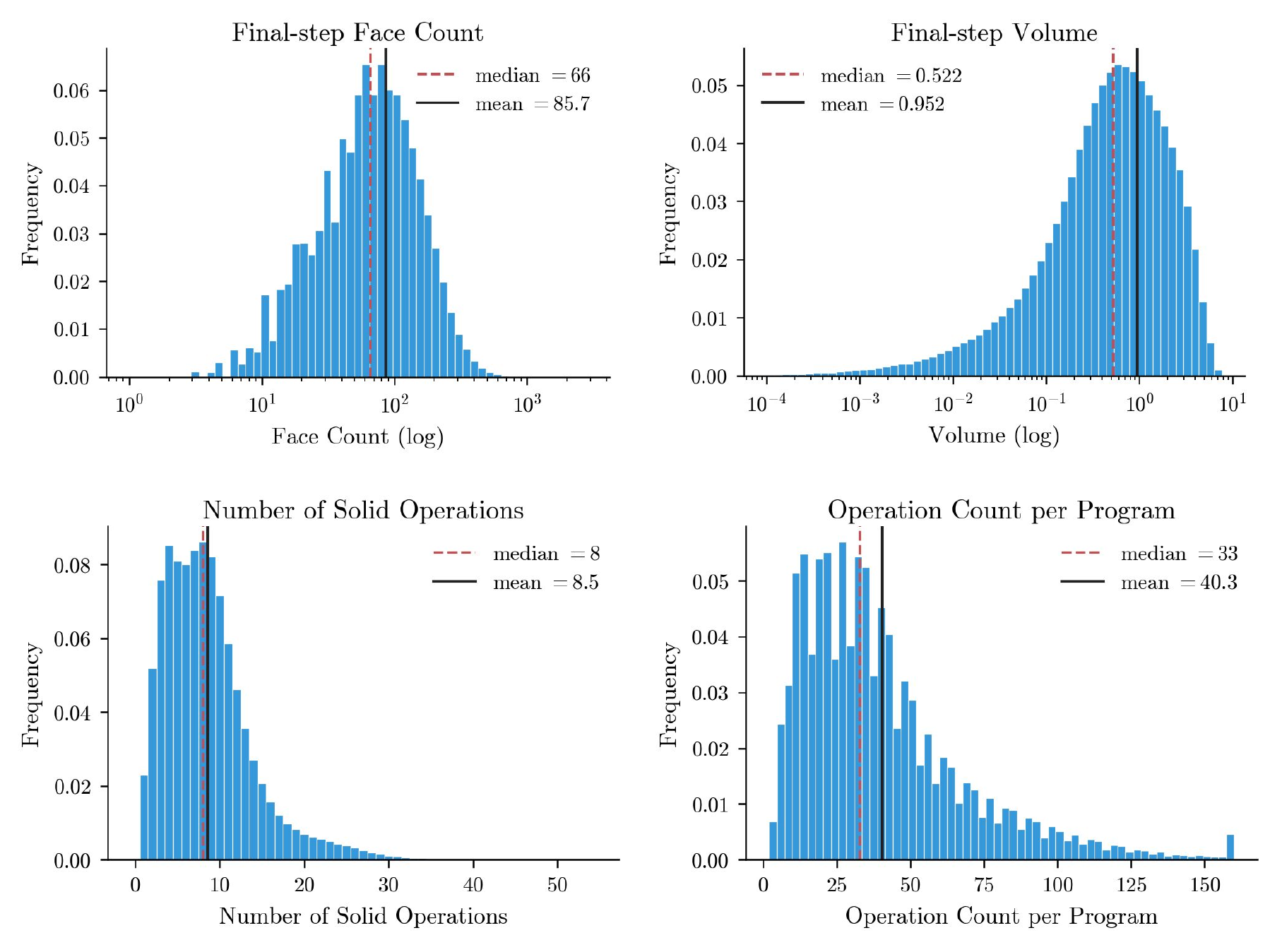}
    \caption{\textbf{ARCADE-1.5M dataset statistics.} 
    We report distributions of face count, volume, solid operations and total operations per program. 
    The dataset spans a wide range of shapes and scales, while exhibiting long-horizon construction sequences and non-trivial solid modeling depth. Operation counts range from 3 to 150+, with a median of 33 and a mean of 40.3. Dashed and solid lines denote median and mean, respectively.}
    \label{fig:dataset-stats}
\end{figure}

\section{Dataset Generation Details}
\label{app:dataset}

\subsection{Sketch Database}
\label{app:sketch_db}

The fundamental source of geometric diversity is a curated database
of 2-D sketch loop primitives derived from real engineering CAD
models.
Each sketch entry is a dictionary containing a list of
\emph{primitives}, where every primitive is one of four types:

\begin{itemize}
  \item \textbf{Line}: a pair of $(x, y)$ endpoints.
  \item \textbf{Arc}: three control points (start, mid, end) defining
        a circular arc, encoded as a \texttt{threePointArc}.
  \item \textbf{Circle}: a centre point and a radius.
  \item \textbf{Spline}: an ordered list of $(x, y)$ control points
        defining a B-spline curve.
\end{itemize}

At runtime, \texttt{CADFeatureTreeGenerator.get\_random\_sketch()}
draws a uniformly random entry from this index and loads it on
demand, with an in-memory LRU cache to avoid repeated disk I/O.

\subsection{Sketch Placement and Scaling}
\label{app:sketch_placement}

Before a sketch is used to build a solid it must be fitted into a
caller-specified 3-D axis-aligned bounding box (\emph{bbox}).
The procedure is:

\begin{enumerate}
  \item \textbf{Compute sketch 2-D bbox.}
        All primitive control points are collected and the tight
        axis-aligned bounding box
        $[s_{x}^{\min}, s_{y}^{\min}, s_{x}^{\max}, s_{y}^{\max}]$
        is computed.

  \item \textbf{Select a face of the target 3-D bbox.}
        The six bbox faces are enumerated with their outward normals
        and local $x$-directions.  One face is chosen uniformly at
        random as the workplane for the sketch.

  \item \textbf{Compute uniform scale and 2-D translation.}
        The sketch is uniformly scaled so that it fills a random
        fraction $f$ of the face's 2-D extent,
        then translated so its centre coincides with the face centre:
        \[
          s = f \cdot \min\!\left(\frac{2T_w}{S_w},\,
                                   \frac{2T_h}{S_h}\right),
          \qquad
          c_x = \tfrac{T_x^{\min}+T_x^{\max}}{2},\;
          c_y = \tfrac{T_y^{\min}+T_y^{\max}}{2}.
        \]

  \item \textbf{Apply transform to primitives.}
        Every control point $(p_x, p_y)$ is mapped to
        $(p_x \cdot s + c_x,\; p_y \cdot s + c_y)$ before being
        written into the generated code.
\end{enumerate}

\subsection{Primitive Code Generation}
\label{app:primitive_factory}

Given a placed-and-scaled sketch and a target 3-D bbox,
\texttt{generate\_cadquery\_code()} emits CadQuery Python code for
one of four volumetric operations.

\textbf{Extrude.}
A named workplane is created from
\texttt{cq.Plane(origin=\ldots, normal=\ldots, xDir=\ldots)},
aligned with the chosen bbox face.
The sketch primitives are emitted in order:
lines as \texttt{.moveTo(x0,y0).lineTo(x1,y1)} (first) or
\texttt{.lineTo(x1,y1)} (subsequent);
arcs as \texttt{.threePointArc((x1,y1),(x2,y2))};
circles as \texttt{.moveTo(cx,cy).circle(r)};
splines as \texttt{.spline([(x0,y0),\ldots])}.
A \texttt{.close()} seals the contour and
\texttt{.extrude(d)} creates the solid, where $d$ fills 85--100\%
of the bbox dimension normal to the face.
Example output:
\begin{verbatim}
plane_0 = cq.Plane(origin=(0,0,0), normal=(0,0,1), xDir=(1,0,0))
sketch_0 = cq.Workplane(plane_0)
sketch_0 = sketch_0.moveTo(-1.2,-0.8).lineTo(1.2,-0.8)
sketch_0 = sketch_0.threePointArc((1.5,0.0),(1.2,0.8))
sketch_0 = sketch_0.lineTo(-1.2,0.8).close()
solid_0  = sketch_0.extrude(2.5)
\end{verbatim}

\textbf{Revolve.}
Workplane and sketch emission are identical to extrude.
After \texttt{.close()}, the solid is created with
\texttt{.revolve(angleDegrees=\(\theta\),
axisStart=(ax,ay), axisEnd=(bx,by))},
where $\theta \in \{90,180,270,360\}$ is chosen uniformly and the
axis is either a line extracted from the sketch primitives (50\%
probability) or an edge of the face bbox.
With 25\% probability a circle replaces the database sketch profile.

\textbf{Sweep (spline path).}
A 2-D spline path is generated by sampling one of ten named
archetypes --- \emph{open}: arc, S-curve, U-curve, L-curve, wave,
diagonal; \emph{closed}: ellipse, rounded rectangle, teardrop,
kidney --- and emitted as:
\begin{verbatim}
path_0 = cq.Workplane('XY').spline([(x0,y0),(x1,y1),...])
\end{verbatim}
(with \texttt{.close()} for closed loops).
The profile cross-section on the start face is emitted identically
to the extrude case, then swept with
\texttt{profile\_0.sweep(path\_0)}.

\textbf{Loft (offset).}
A single workplane is created.
The sketch is drawn $N \in [3,7]$ times at successive
\texttt{.workplane(offset=\(\delta_i\))} steps, each with a
per-profile scale factor that creates a tapered or gradually varying
cross-section.  The chain is closed with \texttt{.loft()}.

\textbf{Loft (transformed).}
The base profile is drawn on the start workplane.
Each subsequent profile is reached with
\texttt{.transformed(offset=(\(\Delta x, \Delta y, \Delta z\)),
rotate=(\(r_x, r_y, r_z\)))},
where offsets follow a 2-D spline spine (same archetypes as sweep)
and rotation angles are drawn from the discrete set
$\{0, 10, 20, 30, 45, 60, 90\}^\circ$.
Profile scale also varies per section.
\texttt{.loft()} is called after all profiles are drawn.

\textbf{Per-primitive fillet/chamfer.}
With 50\% probability, each primitive solid is immediately followed
by \texttt{solid\_N = solid\_N.edges().fillet(r)} or
\texttt{.chamfer(d)}, $r, d \in [0.1, 0.9]$, applied as part of the
primitive's own code block before it enters the Boolean tree.

\subsection{SolidFactory: Bounding-Box-Driven Dispatch}
\label{app:solid_factory}

\texttt{SolidFactory} wraps \texttt{CADFeatureTreeGenerator} and
exposes a uniform bounding-box interface: every public method accepts
$(c_x, c_y, c_z, s_x, s_y, s_z)$ (centre + extents) and returns a
validated CadQuery solid together with the code string that built it.
The internal flow is:

\begin{enumerate}
  \item Convert centre/extents to a bbox dict.
  \item Call the appropriate \texttt{generate\_*} method.  The
        generator selects a face, scales the sketch, and returns a
        feature dict.
  \item Pass the feature dict to \texttt{generate\_cadquery\_code()},
        which produces a Python string.
  \item \texttt{exec()} the string and extract \texttt{final\_result}.
  \item Validate: single body, volume $> 0.01$, B-Rep valid
        (\texttt{BRepCheck\_Analyzer.IsValid()}).
  \item Strip the import header and \texttt{final\_result} line;
        store the remaining code in \texttt{self.\_last\_code} and
        the variable name in \texttt{self.\_last\_var}.
\end{enumerate}

Fallback hierarchy on failure:
sketch-based extrude $\to$ cylinder $\to$ axis-aligned box.

\subsection{ModelBuilderV2: Composing Primitives into Parts}
\label{app:builder}

\texttt{ModelBuilderV2} is a stateful accumulator maintaining:
\texttt{self.solid} (the current CadQuery solid),
\texttt{self.\_code\_lines} (a growing list of code-string
fragments), and \texttt{self.\_model\_var} (the Python variable name
of the current combined solid).

At each step an operation is attempted in memory.
If it produces a valid single-body solid, the operation's code
fragment is appended and the solid updated; otherwise the step is
silently skipped, making the builder failure-tolerant.

\textbf{Base creation.}
\texttt{create\_base()} calls \texttt{SolidFactory} with one of the
four primitive types (or all four in random order until one succeeds)
to create the initial solid, centred at the origin with depth
extending along $+Z$.
\texttt{create\_nested\_base()} builds a solid from an \emph{outer}
polygon profile with one or more \emph{inner} cutout profiles
(simulating perforated cross-sections such as frames or flanged
tubes).

\textbf{Union bosses.}
\texttt{add\_boss()} / \texttt{add\_boss\_ex()} place a new
primitive on a named face (top, bottom, front, back, left, right).
\texttt{\_face\_region()} maps fractional width/height/depth
parameters and a lateral offset $(u, v)$ into a 3-D bbox, with an
outward overlap to anchor the boss to the face surface.
The primitive is generated inside that box and boolean-unioned:
\begin{verbatim}
result = result.union(solid_N)
\end{verbatim}
\texttt{add\_step()} / \texttt{add\_multi\_step()} place wide,
shallow bosses covering progressively smaller face fractions to
build stepped profiles.

\textbf{Intersection constraints for union bosses.}
All boss placement functions pass an \texttt{overlap\_frac}
parameter to \texttt{\_face\_region()}, which pushes the boss bbox
inward by that fraction of the face depth so the boss is geometrically
anchored to the host solid rather than merely touching it.
The overlap fraction is sampled as $[0.05, 0.20]$ for regular bosses
and $[0.05, 0.15]$ for steps.
This is a placement-level geometric constraint (bbox overlap), not a
volumetric ratio check.

\textbf{Boolean cuts.}
\texttt{add\_pocket()} and \texttt{add\_through\_cut()} compute an
inward face region, generate a tool (90\% database-sketch extrude,
5\% cylinder, 5\% rect box), then apply a volumetric intersection
constraint before committing: \texttt{has\_sufficient\_intersection()}
computes
\[
  \frac{V(\text{tool} \cap \text{solid})}{V(\text{tool})} \;\geq\; 0.20
\]
and rejects the cut if the ratio falls below 20\%, preventing tools
that barely clip the surface.
\begin{verbatim}
result = result.cut(solid_N)
\end{verbatim}
Through-cuts use a tool depth of $1.2\times$ the solid's full depth
to guarantee complete penetration.
\texttt{add\_slot()} produces a narrow elongated cut
(aspect ratio $\sim$6:1).
\texttt{add\_nested\_cut()} subtracts a nested outer-with-inner
profile tool.

\textbf{Spatial patterns.}
All three pattern types emit \texttt{for}-loop code.

\emph{Linear pattern} --- equally spaced tools along one face axis:
\begin{verbatim}
for i in range(N):
    pos = base + step * (i + 0.5)
    plane_i = cq.Plane(origin=(pos, cy, z0), ...)
    sketch_i = cq.Workplane(plane_i)
    sketch_i = sketch_i.circle(r)           # or polygon
    pat_tool = sketch_i.extrude(h)
    result   = result.cut(pat_tool)         # or .union()
\end{verbatim}

\emph{Circular pattern} --- $N$ equally-spaced tools on a circle of
radius $\rho$ centred at $(c_u, c_v)$:
\begin{verbatim}
for i in range(N):
    angle = phi0 + 2*math.pi*i/N
    fu = cu + rho*math.cos(angle)
    fv = cv + rho*math.sin(angle)
    ...
\end{verbatim}

\emph{Grid pattern} --- doubly-nested loop over rows $\times$ cols:
\begin{verbatim}
for ir in range(rows):
  for ic in range(cols):
    pu = bu + su*(ic+0.5)
    pv = bv + sv*(ir+0.5)
    ...
\end{verbatim}

Each pattern commits only if $\geq 2$ individual tool applications
succeed.

\textbf{Mirror.}
\texttt{add\_mirror()} reflects the solid across XY, XZ, or YZ and
unions the reflection.
With 50\% probability an off-centre mirror plane is used via a
\texttt{basePointVector}.

\textbf{Shell.}
\texttt{add\_shell()} removes a selected face and hollows the solid
to a specified wall thickness:
\begin{verbatim}
result = result.faces('>Z').shell(-t)
\end{verbatim}

\textbf{Intermediate fillet/chamfer.}
After union and cut operations, a 25--50\% probability inline fillet
or chamfer is applied to the sub-solid before the next operation,
simulating mid-process edge-breaking in machining.
Radius/distance is 1--4\% of the current solid's minimum dimension.

\textbf{Final fillet/chamfer.}
\texttt{apply\_fillet()} and \texttt{apply\_chamfer()} are called at
the end of a strategy, with radius/distance drawn as 2--20\% of the
minimum bounding-box dimension.
On failure a smaller value (50\%, 25\%, or 10\% of original) is
tried automatically.

\subsection{Code Assembly and Post-Processing}
\label{app:codegen}

After construction, \texttt{get\_code()} assembles the final
self-contained Python script as follows:

\begin{enumerate}
  \item Prepend \texttt{import cadquery as cq; import math}.
  \item Join all \texttt{\_code\_lines} fragments in sequence.
  \item Append \texttt{final\_result = result} and
        \texttt{cq.exporters.export(final\_result, 'model.step')}.
  \item \textbf{Variable renumbering} (\texttt{\_renumber\_vars}):
        \texttt{plane\_N}, \texttt{sketch\_N}, \texttt{solid\_N},
        \texttt{loft\_N}, \texttt{profile\_N}, \texttt{path\_N}
        indices are made contiguous.
        Gaps arise because the factory counter increments even on
        failed attempts.
  \item \textbf{Variable substitution}
        (\texttt{\_substitute\_result\_var}):
        \texttt{solid\_0} is renamed to \texttt{result};
        intermediate bare \texttt{result} assignments are removed.
  \item \textbf{Float rounding} (optional, \texttt{--round-floats}):
        every float literal is rounded to three decimal places via
        regex substitution, reducing token count for language-model
        training.
  \item \textbf{Collinear point removal}
        (\texttt{\_remove\_collinear\_points}):
        any \texttt{lineTo} point that lies between its predecessor
        and successor on the same line, or duplicates either
        endpoint, is eliminated.
        Chained \texttt{.moveTo().lineTo()} calls are first split
        onto separate lines so that rounding-induced duplicates are
        also caught.
\end{enumerate}

\subsection{Strategy Library}
\label{app:strategies}

Ninety-four named strategy functions orchestrate the builder,
organised into seven tiers by complexity and intent:

\begin{description}
  \item[Tier 1 --- Single-primitive (1--4).]
        One primitive type forms the entire body
        (\texttt{strategy\_extruded\_block},
        \texttt{strategy\_revolved\_body},
        \texttt{strategy\_lofted\_body},
        \texttt{strategy\_swept\_body}).

  \item[Tier 2 --- Mid-level composition (5--20).]
        A primary primitive receives 2--6 additional features
        (bosses, pockets, patterns, mirror, shell).
        Examples: \texttt{strategy\_symmetric\_extruded}
        (extrude $\to$ features on one half $\to$ mirror),
        \texttt{strategy\_lofted\_enclosure}
        (loft $\to$ shell $\to$ mounting bosses).

  \item[Tier 3 --- Higher-level composition (21--30).]
        10--18-feature sequences mimicking specific mechanical
        part families.

  \item[Tier 4 --- Union-focused (31--40).]
        Emphasise additive material: multi-scale boss clusters,
        tall fin arrays, nested cylinder rings, pancake stacks,
        hub-and-arm assemblies.

  \item[Tier 5 --- Mega-composite (41--48).]
        Combine two or more mid-level sub-strategies within one
        builder instance, yielding 12--25 features.

  \item[Tier 6 --- Long-sequence (49--72).]
        Detail-pass and nested-sketch strategies targeting 20--50+
        features: surface nubs, micro-pockets, screw-boss panels,
        multi-hole plates, nested-sketch towers.

  \item[Tier 7 --- Operation-subset boost (73--94).]
        Each strategy restricts itself to underrepresented operations
        to rebalance the operation-type distribution:
        revolve-only, loft-only, sweep-only, cut-dominant,
        shell+fillet+chamfer focus, circular-pattern-only,
        mirror-heavy, sweep/loft alternating, grid/slot textured,
        revolve-cut lathe, circular-union crown, linear-union array.
\end{description}

\subsection{Parallel Generation and Double Validation}
\label{app:parallel}

\textbf{Worker architecture.}
Generation is parallelised across $W$ workers (default: all CPU
cores).
Each worker is an isolated subprocess spawned via
\texttt{subprocess.Popen} rather than
\texttt{multiprocessing.Process}, because forking from a
\texttt{ThreadPoolExecutor} thread is unsafe with OpenCASCADE's
global C\texttt{++} state.
A per-model timeout (default: 120\,s) sends \texttt{SIGKILL} to
workers that hang inside the kernel.

\textbf{Double validation.}
Each candidate model undergoes two independent validity checks:

\begin{enumerate}
  \item \textbf{In-memory validation.}
        The builder solid must satisfy:
        (a)~exactly one connected body (\texttt{count\_solids}=1),
        (b)~volume $> 0.01$,
        (c)~\texttt{BRepCheck\_Analyzer.IsValid()} returns
        \texttt{True}.

  \item \textbf{Script re-execution validation.}
        The generated \texttt{.py} script is independently
        \texttt{exec()}-ed in a fresh namespace; the resulting solid
        must pass the same three checks.
        The STEP file is exported from \emph{this} re-executed solid,
        guaranteeing that the stored script and the stored geometry
        are identical.
\end{enumerate}

\textbf{File management.}
Successful models are stored in chunk subdirectories of 10\,000
files each (\texttt{0000/}, \texttt{0001/}, \ldots).
Partial files from failed workers are cleaned up before committing.
Completed runs write \texttt{metadata.json} (strategy, random seed,
geometric scale, full feature history with parameters,
axis-aligned bounding box, volume) and
\texttt{strategy\_stats.json} (per-strategy
success/invalid/timeout/error counts).

\subsection{Dataset Characteristics}
\label{app:stats}

\textbf{Artifact types per model.}
\begin{itemize}
  \item \texttt{model\_XXXXX.py}: self-contained CadQuery script
        (ground-truth program).
  \item \texttt{model\_XXXXX.step}: STEP B-Rep geometry (exported
        from the re-executed script).
  \item \texttt{model\_XXXXX.stl}: triangulated mesh (optional).
\end{itemize}

\textbf{Program characteristics.}
Scripts consist of sequences of CadQuery API calls with explicit
workplane definitions (\texttt{cq.Plane}), sketch primitives
(\texttt{moveTo}, \texttt{lineTo}, \texttt{threePointArc},
\texttt{circle}, \texttt{spline}, \texttt{close}), volumetric
operations (\texttt{extrude}, \texttt{revolve}, \texttt{loft},
\texttt{sweep}), Boolean combinators (\texttt{union}, \texttt{cut}), topology operations (\texttt{mirror}, \texttt{shell}), finishing operations (\texttt{fillet}, \texttt{chamfer}), and \texttt{for}-loop spatial patterns. Float literals are rounded to three decimal
places.

\newpage

\section{Policy Architecture and Training Details}
\label{app:policy_details}

\subsection{Inputs}

At each step, the model receives:
\begin{itemize}
    \item \textbf{Target geometry} $m$: 1024-point cloud sampled from the input mesh.
    \item \textbf{Intermediate geometry} $S_{t-1}$: 1024-point cloud from the current reconstruction.
    \item \textbf{Geometry metadata}: bounding box center and scale of $S_{t-1}$ in the target frame.
    \item \textbf{Program context}: the partial CAD program encoded as text tokens.
\end{itemize}

\subsection{Point Cloud Encoding}

Both target and intermediate point clouds are encoded using a Point-MAE encoder:
\begin{itemize}
    \item Farthest Point Sampling: $1024 \rightarrow 64$ groups
    \item KNN grouping ($k=32$)
    \item PointNet-based per-group encoder
    \item Transformer encoder (12 layers, partial fine-tuning)
\end{itemize}

This produces 64 tokens per point cloud, which are projected to the LLM embedding space.

\subsection{Geometry Conditioning}

A lightweight MLP encodes geometric metadata (center + scale) into a token that preserves global pose information lost during normalization.

\subsection{Multimodal Token Fusion}

Geometric tokens are injected into the language model by replacing reserved placeholder tokens in the input sequence. This yields a unified embedding stream combining:
\begin{itemize}
    \item target geometry tokens
    \item intermediate geometry tokens
    \item geometric metadata
    \item CAD program text
\end{itemize}

\subsection{Language Model}

We use Qwen2.5-Coder-1.5B as the backbone. The model autoregressively predicts the next construction action as CadQuery code, including continuous parameters represented as digit sequences. Prompt: ``Generate CAD operation $\{\text{op}_{i+1}\}$ for this shape:''

\subsection{Decoding and Execution}

At inference time:
\begin{itemize}
    \item The model generates one construction action at a time
    \item The action is executed to produce updated geometry
    \item The resulting geometry is re-encoded for the next step
\end{itemize}

\subsection{Training Setup}

\begin{itemize}
    \item Cross-entropy loss over operation tokens
    \item Full fine-tuning of the language model
    \item Separate learning rates for point cloud encoder (0.00005) and LLM (0.0002)
    \item 1000 learning-rate warmup steps followed by cosine decay
    \item Training on $\sim$12.5M state-action transitions
    \item Training performed on 8 NVIDIA H100 GPUs for 12 hours, batch size of 32, for 100k steps
\end{itemize}

\subsection{Additional StepCAD Details}
\label{app:stepcad_details}

\subsubsection{Problem Formulation}

Let $m \in \mathcal{M}$ denote an input mesh, and let $Q_m \in \mathbb{R}^{N \times 3}$ denote the corresponding sampled point cloud. A CAD program is represented as
\[
p=(a_1,\dots,a_T),
\]
where $a_t \in \mathcal{A}$ are executable construction actions and $S(p)$ is the resulting geometry. The reconstruction objective is:
\[
p^*=\arg\max_{p\in\mathcal{P}} \mathrm{IoU}(S(p),m),
\]
where $\mathcal{P}$ denotes the space of valid CAD programs.

Direct optimization over $\mathcal{P}$ is combinatorial and highly non-convex due to the discrete, sequential, and structured nature of CAD construction programs.

\subsubsection{Autoregressive Policy Formulation}

We model CAD generation as a sequential decision process. Let $Q_t$ denote the point cloud sampled from the intermediate geometry $S_t$. At each step, the policy predicts:
\[
\pi_\theta(a_t \mid Q_{t-1}, Q_m, a_{<t}),
\]
which induces a distribution over complete programs:
\[
\pi_\theta(p)=
\prod_{t=1}^{T}
\pi_\theta(a_t \mid Q_{t-1},Q_m,a_{<t}).
\]

Target and intermediate geometries are encoded using a shared Point-MAE encoder:
\[
z_m=f_\phi(Q_m), \qquad
z_{t-1}=f_\phi(Q_{t-1}),
\]
which are injected into the transformer as conditioning tokens.

The policy is instantiated using Qwen2.5-Coder-1.5B operating jointly over geometry embeddings and autoregressive program tokens. The next construction action is decoded as:
\[
a_t=
\arg\max_a
\pi_\theta(a \mid z_m,z_{t-1},a_{<t}),
\]
or sampled from the predictive distribution during inference-time sampling.

\subsubsection{Dense State-Action Supervision}

The policy is trained using teacher forcing over dense state-action transitions extracted from ARCADE-1.5M:
\[
\mathcal{L}(\theta)=
-
\mathbb{E}_{(Q_{t-1},Q_m,a_t)}
\left[
\log
\pi_\theta(a_t \mid Q_{t-1},Q_m,a_{<t})
\right].
\]

Unlike supervision only on final programs, ARCADE-1.5M provides intermediate geometry after every solid-operation action, yielding dense trajectory-level supervision over approximately 12.5M construction states. This improves long-horizon robustness and geometric consistency.

\subsubsection{Closed-Loop Rollout Procedure}

At inference time, generation alternates between autoregressive prediction and CAD execution. Starting from an empty geometry $S_0=\emptyset$, at each step:
\begin{enumerate}[leftmargin=1.4em,itemsep=2pt,topsep=2pt]
    \item The current intermediate geometry $Q_{t-1}$ and target geometry $Q_m$ are encoded.
    
    \item The next construction action is sampled from
    \[
    a_t \sim
    \pi_\theta(\cdot \mid Q_{t-1},Q_m,a_{<t}),
    \]
    using inference-time sampling.
    
    \item The updated program is executed using CadQuery to obtain the next geometry:
    \[
    S_t=\mathrm{Exec}(a_1,\dots,a_t).
    \]
    
    \item A new point cloud $Q_t$ is sampled from $S_t$ and fed back into the policy.
    
    \item Reconstruction quality is evaluated using:
    \[
    r_t=\mathrm{IoU}(S_t,m).
    \]
\end{enumerate}

Generation terminates once reconstruction quality plateaus (IoU improvement below $0.001$), an end-of-program token is emitted, or $T_{\max}=12$ construction actions have been emitted. This is an action cap, not a cap of 12 sketch/solid/Boolean operations or 12 API calls. ARCADE-1.5M averages 40.3 counted operations but only 8.5 solid operations per program, so programs containing more than 40 counted operations may fit within 12 actions. This does not guarantee recovery of shapes whose required construction exceeds the action budget.

This closed-loop execution is important because the policy conditions on the geometry actually produced so far rather than on teacher-forced intermediate states.

\subsubsection{Motivation for Program Refinement}

The learned policy efficiently captures the global structure of target shapes. However, due to the autoregressive and discretized nature of generation, small structural or parametric errors can accumulate over long construction sequences, motivating the geometry-aware refinement stage described in the main paper.

\newpage

\section{Geometry-Aware Program Optimization Details}
\label{app:treesearch}

\subsection{Beam Search Procedure}

Given a seed program $\hat{p} = (a_1,\dots,a_T)$ sampled from the policy $\pi_\theta$, we perform local optimization in program space:
\[
p^* = \arg\max_{p \in \mathcal{N}(\hat{p})} \ \mathrm{IoU}(S(p), m),
\]
where $S(p)$ denotes the geometry produced by executing program $p$ and $\mathcal{N}(\hat{p})$ is a neighborhood induced by local program edits.

Searching all edits jointly is exponential in the sequence length $T$. Instead, we exploit the sequential structure of CAD programs and perform a geometry-aware beam search over operations in execution order. At depth $t$, each candidate in the current beam is expanded by applying local edits to operation $a_t$, yielding:
\[
\mathcal{B}_t =
\mathrm{TopK}_K
\left\{
p' \in \mathrm{Expand}(p,t)
:\ p \in \mathcal{B}_{t-1}
\right\},
\]
where candidates are ranked by:
\[
\mathrm{score}(p)=\mathrm{IoU}(S(p),m).
\]

We use beam width $K=2$ and $M=4$ local edits per expansion. At depth $t=1$, the initial program expands into four candidates and the best two are retained. At subsequent depths, each surviving candidate produces four neighbors, resulting in eight evaluated candidates per depth.

The final refined program is selected as:
\[
p^* =
\arg\max_{p \in \mathcal{B}_T}
\mathrm{IoU}(S(p),m).
\]

\begin{algorithm}[t]
\caption{Geometry-aware tree search for CAD program refinement.}
\label{alg:treesearch}
\small
\begin{algorithmic}[1]
\Require Target mesh $m$, seed program $\hat{p}=(a_1,\dots,a_T)$, beam width $K$, local edits $\{e_1,\dots,e_M\}$
\State $\mathcal{B}_0 \gets \{\hat{p}\}$
\For{$t=1$ to $T$}
    \State $\mathcal{C} \gets \emptyset$
    \For{$p \in \mathcal{B}_{t-1}$}
        \For{$j=1$ to $M$}
            \State $p' \gets e_j(p,t)$
            \State Execute $p'$ with CadQuery
            \State $s(p') \gets \mathrm{IoU}(S(p'),m)$
            \State $\mathcal{C} \gets \mathcal{C} \cup \{(p',s(p'))\}$
        \EndFor
    \EndFor
    \State $\mathcal{B}_t \gets$ top-$K$ candidates in $\mathcal{C}$ ranked by $s(\cdot)$
\EndFor
\State \Return $\arg\max_{p \in \mathcal{B}_T}\mathrm{IoU}(S(p),m)$
\end{algorithmic}
\end{algorithm}

\subsection{Continuous Parameter Refinement}

The continuous refinement operator optimizes the scalar parameters of the current operation without modifying program structure. This includes extrusion depths, radii, sketch coordinates, offsets, and angles.

For an operation parameter vector
\[
v \in \mathbb{R}^d,
\]
we define a bounded local search region:
\[
[v_i-r_i,\ v_i+r_i],
\qquad
r_i = \min\!\left(
\max(0.20|v_i|,\ 0.005),
0.30
\right).
\]

Optimization is performed using Bayesian Optimization with a Gaussian-Process surrogate and Expected-Improvement acquisition. We initialize the search with $6$ random samples followed by $10$ BO iterations using batches of $4$ candidates each. For fillet radii and chamfer lengths, we instead use the fixed interval [0.01, 0.5] in normalized coordinate units.

The refined parameters are accepted only if the resulting geometry improves IoU relative to the original operation.

\subsection{Sketch Extraction and Replacement}

The sketch modification operator replaces the sketch associated with the current operation using geometry extracted directly from the target mesh.

Given the operation working plane $(O,\vec n,\vec x)$, we intersect the target mesh with the plane using \texttt{trimesh.section}. The resulting curves are projected into the local 2D coordinate system $(\vec x,\vec y)$ and converted into closed polygons. The largest valid closed loop is selected.

The polygon is simplified using the Ramer--Douglas--Peucker algorithm, after which a primitive-fitting procedure selects among:
\begin{itemize}[leftmargin=1.2em,itemsep=2pt]
    \item full circles,
    \item line-and-arc decompositions,
    \item raw polygonal approximations.
\end{itemize}

The reconstructed sketch is then serialized back into CadQuery sketch commands:
\begin{verbatim}
Workplane(...).moveTo(...).lineTo(...)
.threePointArc(...).close()
\end{verbatim}

The original operation sketch is replaced while preserving the operation type and extrusion depth.

This edit is skipped if the plane does not intersect the target mesh or only open curves are produced.

\subsection{Operation Skipping}

The operation-skipping edit removes the current operation entirely:
\[
p' = p \setminus a_t.
\]

The modified program is executed and accepted only if:
\[
\mathrm{IoU}(S(p'),m)
>
\mathrm{IoU}(S(p),m).
\]

This edit is particularly effective for removing redundant operations, degenerate sketches, or erroneous boolean cuts introduced by the autoregressive policy.

\subsection{Complement-Cut Optimization}

The complement-cut operator removes overshoot regions by generating cut-extrudes from the boolean difference between the generated geometry and the target mesh.

Let:
\[
G = S(p),
\qquad
I = m.
\]
We compute the overshoot region:
\[
D = G \setminus I
\]
using manifold3d boolean operations.

Planar faces of $D$ are clustered by normal direction and plane offset using:
\begin{itemize}[leftmargin=1.2em,itemsep=2pt]
    \item angular tolerance: $2^\circ$,
    \item offset tolerance: $10^{-3}$.
\end{itemize}

Clusters with total area below $0.0025$ are discarded.

For each planar cluster, boundary loops are projected into the local 2D plane and converted into sketches using the same primitive-fitting procedure as sketch replacement. Candidate cut-extrudes are generated by grid-searching extrusion depths:
\[
\{0.05,0.10,0.15,0.20,0.30,0.40,0.60,0.80,1.20,2.00\}.
\]

The best cut operation is appended to the program using:
\[
\texttt{combine='cut'}.
\]

We allow up to two complement-cut rounds, since removing one overshoot region can expose additional dominant errors.

\section{Additional Evaluation: Fidelity, Search Transfer, and Compactness}
\label{app:rebuttal-evaluation}

\subsection{High-IoU Reconstruction Rates}
We report the percentage of evaluated CADBench samples reaching each IoU threshold. High-threshold success measures volumetric fidelity without requiring program correspondence. It does not imply that all dimensions or small features are correct, because global volume overlap can be insensitive to localized errors.
\begin{table}[htbp]
\centering\small
\caption{\textbf{High-IoU success rates on CADBench (\%).} Thresholds assess volumetric reconstruction fidelity, not direct parameter accuracy.}
\label{tab:high-iou}
\begin{tabular}{lrrrr}
\toprule
Method & $\geq0.80$ & $\geq0.90$ & $\geq0.95$ & $\geq0.99$ \\
\midrule
CAD-Recode & 43.8 & 30.6 & 19.5 & 3.9 \\
Cadrille & 44.1 & 30.2 & 18.7 & 3.4 \\
CADEvolve & 45.9 & 32.3 & 21.7 & 5.6 \\
CADReasoner & 49.4 & 38.4 & 27.2 & 7.4 \\
StepCAD & \textbf{71.1} & \textbf{55.3} & \textbf{39.9} & \textbf{15.3} \\
\bottomrule
\end{tabular}
\end{table}

\subsection{Transfer to Independently Generated Programs}
The refinement stage uses the target mesh and an executable seed program, with no policy weights or learned policy features. On 1,200 CADBench samples, applying it to CAD-Recode output increases mean IoU by $0.148$. This demonstrates transfer to another generator's compatible programs; it does not establish support for arbitrary CAD languages or all program structures. The subset results are separate from the full-benchmark results in Table~\ref{tab:high-iou}.
\begin{table}[htbp]
\centering\small
\caption{\textbf{Search applied to CAD-Recode programs} on the same 1,200-sample CADBench subset, without retraining.}
\label{tab:search-transfer}
\begin{tabular}{lrr}
\toprule
Metric & CAD-Recode seed & With StepCAD search \\
\midrule
Mean IoU & 0.597 & \textbf{0.745} \\
IoU $\geq0.80$ (\%) & 45.8 & \textbf{63.4} \\
IoU $\geq0.90$ (\%) & 29.2 & \textbf{45.8} \\
IoU $\geq0.95$ (\%) & 17.1 & \textbf{32.6} \\
\bottomrule
\end{tabular}
\end{table}

\subsection{Entity-Level Geometric Accuracy}
On CADBench datasets with reference B-reps, we additionally evaluate vertex, edge, and face accuracy using the definitions and tolerance of Pointer-CAD v2~\cite{qi2026pointerv2}. StepCAD has the highest reported accuracy at all three levels (Table~\ref{tab:entity-accuracy}). These metrics provide information not captured by global IoU. Their interpretation depends on the matching protocol and reference decomposition: different valid histories can yield similar geometry with different surface representations or subdivisions. This limits their use as a unique measure of construction-history correctness, but does not make entity-level geometry uninformative.
\begin{table}[htbp]
\centering\small
\caption{\textbf{Entity-level accuracy} on CADBench datasets with reference B-reps. Values are the accuracy scores reported in the additional evaluation.}
\label{tab:entity-accuracy}
\begin{tabular}{lrrr}
\toprule
Method & Vertex $\uparrow$ & Edge $\uparrow$ & Face $\uparrow$ \\
\midrule
CAD-Recode & 11.4 & 6.3 & 4.1 \\
CADEvolve & 15.7 & 9.7 & 6.4 \\
StepCAD & \textbf{20.8} & \textbf{12.6} & \textbf{8.1} \\
\bottomrule
\end{tabular}
\end{table}

\subsection{Program Size and Reconstruction Quality}
Table~\ref{tab:program-size} reports mean total counted operations together with IoU. The policy alone produces the shortest programs among the listed methods while exceeding their reported IoU. Search increases mean size from 16.2 to 31.7 operations as IoU improves from approximately $0.68$ to $0.82$. Thus, refinement trades program length for fidelity; it does not uniformly simplify the construction history. Operation counts characterize compactness but do not establish constraint preservation, semantic correctness, or successful downstream editing.
\begin{table}[htbp]
\centering\small
\caption{\textbf{Program size and fidelity on CADBench.} Counts are total operations rather than policy actions; IoU is reported at the precision provided by this analysis.}
\label{tab:program-size}
\begin{tabular}{lrr}
\toprule
Method & Mean operations & IoU $\uparrow$ \\
\midrule
$\Pi$-StepCAD & 16.2 & 0.68 \\
CADEvolve & 19.9 & 0.57 \\
Cadrille & 23.5 & 0.58 \\
CAD-Recode & 25.5 & 0.57 \\
StepCAD & 31.7 & 0.82 \\
\bottomrule
\end{tabular}
\end{table}

\subsection{Conciseness Relative to Reference Programs}
Following the motivation of Fusion 360 Gallery~\cite{willis2021fusion}, we compare recovered sequence length with reference sequence length on DeepCAD. For recovered and reference sequences $A$ and $\hat A$, respectively, the ratio is $|A|/|\hat A|$, using consistent counting units. Table~\ref{tab:conciseness} reports the supplied analysis stratified by B-rep face-count complexity. StepCAD has an overall ratio of $0.91$ and IoU of $0.89$. A smaller ratio is not necessarily better: omitted geometry can shorten a program, so conciseness must be interpreted jointly with reconstruction accuracy. These results are not restricted to exact reconstructions.
\begin{table}[htbp]
\centering\small
\caption{\textbf{Conciseness on DeepCAD.} Recovered/reference sequence-length ratios and overall IoU. Lower ratios alone do not imply better reconstruction or editability.}
\label{tab:conciseness}
\begin{tabular}{lrrrrr}
\toprule
Method & Easy & Medium & Hard & Overall & IoU \\
\midrule
CAD-Recode & 0.99 & 0.99 & 0.71 & 0.90 & 0.79 \\
CADEvolve & 1.02 & 0.82 & 0.53 & 0.79 & 0.74 \\
$\Pi$-StepCAD & 0.95 & 0.87 & 0.57 & 0.80 & 0.82 \\
StepCAD & 1.05 & 0.94 & 0.73 & 0.91 & 0.89 \\
\bottomrule
\end{tabular}
\end{table}

\section*{Assets and licenses}
We evaluate on CADBench, which aggregates data from publicly available datasets (e.g., ABC, Fusion 360, Objaverse). We follow the respective licenses and terms of use of the original sources. We also use pretrained models (Qwen2.5-Coder, Point-MAE) and CadQuery (Apache 2.0), in accordance with their licenses.

\newpage
\section*{NeurIPS Paper Checklist}

\begin{enumerate}

\item {\bf Claims}
    \item[] Question: Do the main claims made in the abstract and introduction accurately reflect the paper's contributions and scope?
    \item[] Answer: \answerYes{}
    \item[] Justification: The claims in the abstract and introduction accurately reflect the paper’s contributions, methodology, and experimental results.

\item {\bf Limitations}
    \item[] Question: Does the paper discuss the limitations of the work performed by the authors?
    \item[] Answer: \answerYes{}
    \item[] Justification: The limitations of the work are mentioned in section \ref{sec:limitations}.

\item {\bf Theory assumptions and proofs}
    \item[] Question: For each theoretical result, does the paper provide the full set of assumptions and a complete (and correct) proof?
    \item[] Answer: \answerNA{}
    \item[] Justification: The paper does not include theoretical results.

    \item {\bf Experimental result reproducibility}
    \item[] Question: Does the paper fully disclose all the information needed to reproduce the main experimental results of the paper to the extent that it affects the main claims and/or conclusions of the paper (regardless of whether the code and data are provided or not)?
    \item[] Answer: \answerYes{}
    \item[] Justification: All the information needed to reproduce the main experimental results of the paper is available in the main paper and appendices.

\item {\bf Open access to data and code}
    \item[] Question: Does the paper provide open access to the data and code, with sufficient instructions to faithfully reproduce the main experimental results, as described in supplemental material?
    \item[] Answer: \answerNo{}
    \item[] Justification: We plan to release the code and dataset upon acceptance. The paper and supplementary material provide detailed implementation, training, and evaluation details intended to support reproducibility of the proposed approach and experiments.

\item {\bf Experimental setting/details}
    \item[] Question: Does the paper specify all the training and test details (e.g., data splits, hyperparameters, how they were chosen, type of optimizer) necessary to understand the results?
    \item[] Answer: \answerYes{} 
    \item[] Justification: The full details are provided in the appendix.

\item {\bf Experiment statistical significance}
    \item[] Question: Does the paper report error bars suitably and correctly defined or other appropriate information about the statistical significance of the experiments?
    \item[] Answer: \answerNo{} 
    \item[] Justification: We do not include statistical confidence intervals or error bars, primarily due to the high computational cost.

\item {\bf Experiments compute resources}
    \item[] Question: For each experiment, does the paper provide sufficient information on the computer resources (type of compute workers, memory, time of execution) needed to reproduce the experiments?
    \item[] Answer: \answerYes{} 
    \item[] Justification: The full details are provided in the appendix.
    
\item {\bf Code of ethics}
    \item[] Question: Does the research conducted in the paper conform, in every respect, with the NeurIPS Code of Ethics \url{https://neurips.cc/public/EthicsGuidelines}?
    \item[] Answer: \answerYes{} 
    \item[] Justification: Our research complies with the NeurIPS Code of Ethics. The dataset is synthetic and poses no identifiable human data or privacy risks.

\item {\bf Broader impacts}
    \item[] Question: Does the paper discuss both potential positive societal impacts and negative societal impacts of the work performed?
    \item[] Answer: \answerYes{} 
    \item[] Justification: We discuss societal impacts in Section \ref{sec:limitations}.
    
\item {\bf Safeguards}
    \item[] Question: Does the paper describe safeguards that have been put in place for responsible release of data or models that have a high risk for misuse (e.g., pre-trained language models, image generators, or scraped datasets)?
    \item[] Answer: \answerNA{} 
    \item[] Justification: The paper poses no such risks.

\item {\bf Licenses for existing assets}
    \item[] Question: Are the creators or original owners of assets (e.g., code, data, models), used in the paper, properly credited and are the license and terms of use explicitly mentioned and properly respected?
    \item[] Answer: \answerYes{}
    \item[] Justification: All external datasets, models, and software used in this work are properly cited, and their corresponding licenses and terms of use are respected and referenced where applicable.

\item {\bf New assets}
    \item[] Question: Are new assets introduced in the paper well documented and is the documentation provided alongside the assets?
    \item[] Answer: \answerYes{}
    \item[] Justification: The paper and supplementary material document the proposed dataset, model, and evaluation pipeline, including details on generation, training, and limitations. We plan to release the assets and accompanying documentation upon acceptance.

\item {\bf Crowdsourcing and research with human subjects}
    \item[] Question: For crowdsourcing experiments and research with human subjects, does the paper include the full text of instructions given to participants and screenshots, if applicable, as well as details about compensation (if any)? 
    \item[] Answer: \answerNA{} 
    \item[] Justification: The paper does not involve crowdsourcing nor research with human subjects.

\item {\bf Institutional review board (IRB) approvals or equivalent for research with human subjects}
    \item[] Question: Does the paper describe potential risks incurred by study participants, whether such risks were disclosed to the subjects, and whether Institutional Review Board (IRB) approvals (or an equivalent approval/review based on the requirements of your country or institution) were obtained?
    \item[] Answer: \answerNA{} 
    \item[] Justification: The paper does not involve crowdsourcing nor research with human subjects.

\item {\bf Declaration of LLM usage}
    \item[] Question: Does the paper describe the usage of LLMs if it is an important, original, or non-standard component of the core methods in this research? Note that if the LLM is used only for writing, editing, or formatting purposes and does \emph{not} impact the core methodology, scientific rigor, or originality of the research, declaration is not required.
    \item[] Answer: \answerNA{} 
    \item[] Justification: The core method development in this research does not involve LLMs as any important, original, or non-standard components.

\end{enumerate}

\end{document}